\documentclass{article} 
\usepackage{sty/iclr2022_conference,times}
\usepackage{sty/mystyle}

\usepackage{amsmath,amsfonts,bm}

\def\1{\bm{1}}

\def\vg{{\bm{g}}}

\def\vx{{\bm{x}}}
\def\vy{{\bm{y}}}
\def\vz{{\bm{z}}}

\def\mI{{\bm{I}}}

\def\mX{{\bm{X}}}

\DeclareMathAlphabet{\mathsfit}{\encodingdefault}{\sfdefault}{m}{sl}
\SetMathAlphabet{\mathsfit}{bold}{\encodingdefault}{\sfdefault}{bx}{n}

\newcommand{\E}{\mathbb{E}}

\DeclareMathOperator*{\argmax}{arg\,max}

\newacronym{cnn}{CNN}{Convolutional Neural Network}

\newacronym{dp}{DP}{Demographic Parity}

\newacronym{elbo}{ELBO}{Evidence Lower Bound}

\newacronym{groupmig}{group-MIG}{Group Mutual Information Gap}

\newacronym{kl}{KL}{Kullback-Leibler}

\newacronym{mc}{MC}{Monte Carlo}
\newacronym{mi}{MI}{Mutual Information}
\newacronym{mig}{MIG}{Mutual Information Gap}
\newacronym{mlp}{MLP}{Multi-Layer Perceptron}

\newacronym{nmi}{NMI}{Normalized Mutual Information}

\newacronym{vae}{VAE}{Variational Autoencoder}

\definecolor{mediumpersianblue}{rgb}{0.0, 0.4, 0.65}
\usepackage{hyperref}
\hypersetup{colorlinks,linkcolor={blue},citecolor={mediumpersianblue},urlcolor={red}} 
\usepackage{url}

\title{Group-disentangled Representation Learning with Weakly-Supervised Regularization}
\iclrfinalcopy
\author{Linh Tran \\
Autodesk AI Lab \\
\& Imperial College London\\
\texttt{linh.tran@autodesk.com} 
\And
Amir Hosein Khasahmadi \\
Autodesk AI Lab \\
\texttt{amir.khasahmadi@autodesk.com} 
\And
Aditya Sanghi \\
Autodesk AI Lab \\
\texttt{aditya.sanghi@autodesk.com} 
\And
Saeid Asgari \\
Autodesk AI Lab \\
\texttt{saeid.asgari.taghanaki@autodesk.com} \\
}

\begin{document}

\maketitle

\begin{abstract}
Learning interpretable and human-controllable representations that uncover factors of variation in data remains an ongoing key challenge in representation learning. We investigate learning group-disentangled representations for groups of factors with weak supervision. Existing techniques to address this challenge merely constrain the approximate posterior by averaging over observations of a shared group. As a result, observations with a common set of variations are encoded to distinct latent representations, reducing their capacity to disentangle and generalize to downstream tasks. In contrast to previous works, we propose GroupVAE, a simple yet effective Kullback-Leibler (KL) divergence-based regularization across shared latent representations to enforce consistent and disentangled representations. We conduct a thorough evaluation and demonstrate that our GroupVAE significantly improves group disentanglement. Further, we demonstrate that learning group-disentangled representations improve upon downstream tasks, including fair classification and 3D shape-related tasks such as reconstruction, classification, and transfer learning, and is competitive to supervised methods. 
\end{abstract}

\section{Introduction}
Decomposing data into disjoint independent factors of variations, i.e., learning disentangled representations, is essential for interpretable and controllable machine learning \citep{bengio2013representation}. 
Recent works have shown that disentangled representation is useful for abstract reasoning \citep{SteenkisteLSB19}, fairness \citep{locatello2019fairness,creager2019flexibly}, reinforcement learning \citep{HigginsPRMBPBBL17} and general predictive performance \citep{locatello2019challenging}.
While there is no consensus on the definition of disentanglement, existing works define it as learning to separate all factors of variation in the data \citep{bengio2013representation}.
According to this definition, altering a single underlying factor of variation should only affect a single factor in the learned representation. However, works in learning disentangled representations \pcite{higgins2016beta,ChenLGD18,locatello2019challenging} have shown that this setting comes with a trade-off between the precision of the representation and the fidelity of the samples.
Therefore, learning precise representations for finer factors, i.e., each factor of variation, may not be practical or desirable.
We deviate from this stringent assumption to learn \textit{group-disentangled representations}, in which a group might include several factors of variation that can co-variate. For instance, groups of interest may be content, style, or background. As a result, a change in one component might affect other variables in a group but not on other groups.

We present \textit{GroupVAE}, a \pacrfull{vae} based framework that leverages weak supervision to learn group-disentangled representations. 
In particular, we use paired observations that always share a group of factors.
Existing group-disentangled approaches \citep{bouchacourt2018multi,hosoya2019group} enforce disentangled group representations by using an average or product of approximate group posteriors.
However, as group representations are dependent on the observations used for the average or product, observations belonging to the same group may not be encoded to the same latent representations. 
We address this inconsistency challenge by incorporating a simple but effective regularization based on the \pacrfull{kl} divergence.
Our idea builds on maximizing the \pacrfull{elbo} of the \pglspl{vae} while minimizing the \pgls{kl} divergence between the latent variables that correspond to the group shared by the paired observations.

In summary, we make the following contributions:
\begin{enumerate}
    \item We propose a way of learning disentangled representations from paired observations that employs \pgls{kl} regularization between the corresponding groups of latent variables. 
    \item We propose \pgls{groupmig}, a mutual information-based metric for evaluating the effectiveness of group disentanglement methods.
    \item Through extensive evaluation, we show that our GroupVAE's effectiveness on a wide range of applications. Our evaluation shows significant improvement for group disentanglement, fair facial attribute classification, and 3D shape-related tasks, including generation, classification, and transfer learning.
\end{enumerate}

\section{Background \& Notation}\label{sec:background}
\begin{figure}[t!]
    \vspace{-1cm}
    \centering
     \subcaptionbox{Model\label{fig:autoencoder}}{
        \includegraphics[width=0.85\linewidth]{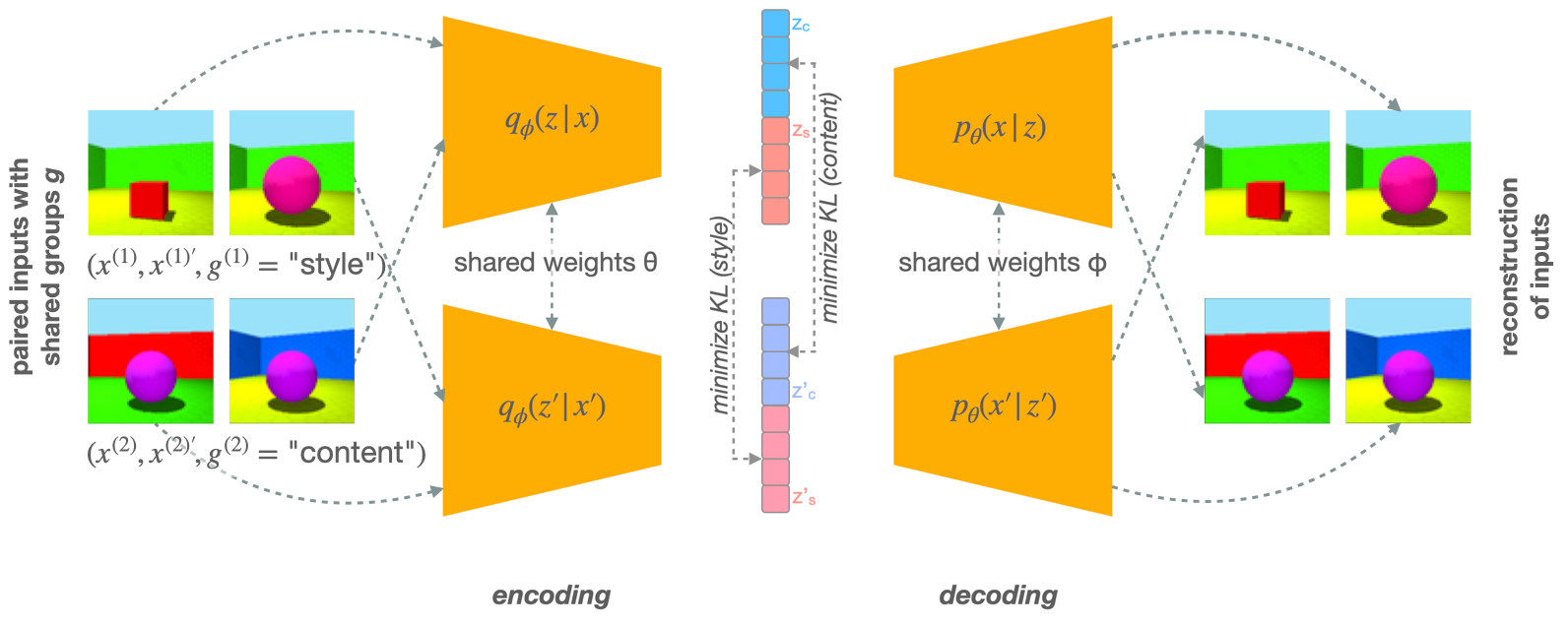}
        \vspace{-0.1cm}
    }\\
    \subcaptionbox{Generative\label{fig:generative}}{
         \resizebox{0.13\linewidth}{!}{
             \begin{tikzpicture}[
    >=stealth, 
    shorten > = 1pt, 
    auto,
    node distance = 2cm, 
    semithick, 
    ]
    \tikzstyle{every state}=[
        draw=black,
        thick,
        fill=white,
        minimum size=8mm
    ]
    \node[state,minimum size=1cm] (zs) at (2, 1.75) {$\vz_{\NonContent}$};
    \node[state,minimum size=1cm] (zg) at (3.5, 1.75) {$\vz_{\Content}$};
    \node[state,fill=gray!30,minimum size=1cm] at (2.75, 0) (x) {$\vx$};
    \draw[->] (zs) -- (x);
    \draw[->] (zg) -- (x);
\end{tikzpicture}
         }
         \vspace{0.95cm}
     }
     \subcaptionbox{Inference\label{fig:inference}}{
         \resizebox{0.25\linewidth}{!}{
             \begin{tikzpicture}[
    >=stealth, 
    shorten > = 1pt, 
    auto,
    node distance = 2cm, 
    semithick, 
    ]
    \tikzstyle{every state}=[
        draw=black,
        thick,
        fill=white,
        minimum size=8mm
    ]
    \node[state,minimum size=1cm] (zs) at (0, 1.1) {$\vz_{\NonContent}$};
    \node[state,minimum size=1cm] (zg) at (1.5, 1.1) {$\vz_{\Content}$};
    \node[state,minimum size=1cm] (zgp) at (3., 1.1) {$\vz'_{\Content}$};
    \node[state,minimum size=1cm] (zsp) at (4.5, 1.1) {$\vz'_{\NonContent}$};
    \node[state,fill=gray!30,minimum size=1cm] at (0.75, 2.85) (x) {$\vx$};
    \node[state,fill=gray!30,minimum size=1cm] at (3.75, 2.85) (xp) {$\vx'$};
    \node at (2.25, -1) {KL};
    \node at (2.25, 2.25) {KL};
    \draw[->] (x) -- (zs);
    \draw[->] (x) -- (zg);
    \draw[->] (xp) -- (zgp);
    \draw[->] (xp) -- (zsp);
    \draw[dashed,<->] (zs.south) to[bend right=60] (zsp.south);
    \draw[dashed,<->] (zg.north) to[bend left=55] (zgp.north);
\end{tikzpicture}
         }
     }
    \vspace{-0.1cm}
    \caption{\textbf{GroupVAE's architecture visualization and graphical model.} We visualize the complete model, including model weights in (a) as well as show the (b) generative and (c) inference parts as graphical models. The model visualization shows two paired inputs, one pair sharing ``style'' and the other sharing ``content''. The \pgls{kl} minimization depends on the group $g$ that is shared. For instance, for input $(\vx^{(1)}, \vx^{'(1)}, g^{(1)}=\textrm{``style''})$, GroupVAE objective only minimizes the \pgls{kl} between the style latent variables. Shaded nodes denote observed quantities in (b) and (c), and unshaded nodes represent unobserved (latent) variables. Dotted arrows represent minimizing the \pgls{kl} divergence between variables during inference.} 
    \label{fig:model_vis}
    \vspace{-0.25cm}
\end{figure}
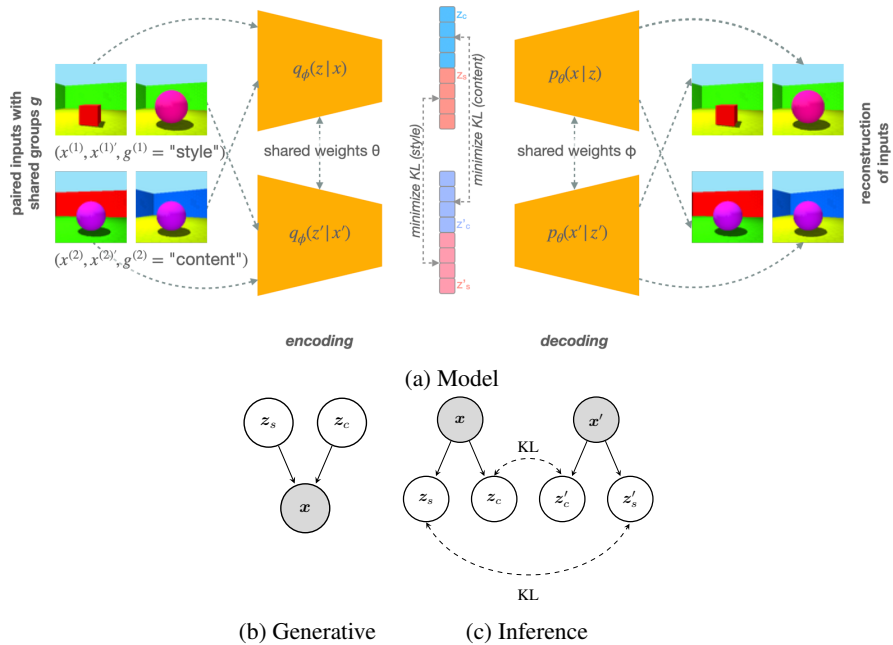

\paragraph{\pacrfull{vae}.} Consider observations $\mX = \{\vx^{(1)}, \ldots, \vx^{(N)} \}, \; \vx^{(i)} \in \mathbb{R}^D$ sampled i.i.d. from distribution $p_{\mX}$ and latent variables $\vz$. A \pacrfull{vae} learns the joint distribution $p(\vx, \vz) = p_{\theta}(\vx | \vz) p(\vz)$ where $p_{\theta}(\vx|\vz)$ is the likelihood function of observations $\vx$ given $\vz$, $\theta$ are the model parameters of $p$ and $p(\vz)$ is the prior of the latent variable $\vz$. \pglspl{vae} are trained to maximize the evidence lower bound (ELBO) on the log-likelihood $\log p(\vx)$. This objective averaged over the empirical distribution is given as 
\begin{align}
    \mathcal{L} = \frac{1}{N} \sum_{i=1}^N \big(\mathbb{E}_{q}[\log p(\vx^{(i)}|\vz)] - \textrm{KL}(q_{\phi}(\vz|\vx^{(i)}) || p(\vz)), \label{eq:elbo}
\end{align}
where $q_{\phi}(\vz|\vx)$ denotes the learned approximate posterior, $\phi$ the variational parameters of $q$ and KL denotes the Kullback-Leibler (KL) divergence.
VAEs \pcite{KingmaW13} are frequently used for learning disentangled representations and serve as the basis of our approach. 

\paragraph{Weakly-supervised group disentanglement.}
We assume the observations $\mathcal{\mX}=\{\vx^{(1)}, \ldots, \vx^{(N)}\}$ 
and the data generating process can be described by
$M$ distinct groups $G=\{g_1,\ldots,g_M\}$.
Each group splits $\mathcal{\mX}$ into disjoint partitions with arbitrary sizes.
Each group consists of non-overlapping sets of factors of variations.
For example, images of 3D shapes \citep{3dshapes18}\footnote{Samples are shown in  Figure~\ref{fig:model_vis}.} can be described through three groups: \textit{shape}\footnote{The group \textit{shape} contains factors such as shape category, shape size and shape color.}, \textit{background}\footnote{The group \textit{background} contains factors such as floor color, wall color.} and \textit{view}. Without loss of generality, we define two groups $g_{\ucContent}$ (\textit{content}) and $g_{\ucNonContent}$ (\textit{style} independent of content) to describe the generative and inference process. We assume having paired observations $(\vx, \vx')$ for training in a weakly-supervised setting.
Each pair of observations shares the same group, i.e., in our case either content $\Content \in g_{\ucContent}$ or style $\NonContent \in g_{\ucNonContent}$. During inference, the exact values for content and style are unknown, but only that $(\vx, \vx')$ share a certain group is known.
For each observation $\vx$, we define two latent variables: $\vz_{\Content}$ for content and $\vz_{\NonContent}$ for style. The goal for group-based disentanglement is that the representation for the same group as close to each other to ensure consistency. 
\section{Learning Group-Disentangled Representations}\label{sec:groupvae}
In the following, we introduce \textit{GroupVAE}, a deep generative model which learns disentangled representations for each group of factors. For simplicity, we limit the formulation of GroupVAE to two groups, content and style, although GroupVAE can be applied to any number of groups.
This section first describes the generative and inference model and then introduces our main contributions -- the \pgls{kl} regularization and inference scheme. We visualized the generative and inference model in Figures~\ref{fig:generative} and~\ref{fig:inference}. 
\paragraph{Inference and generative model.} Our model uses paired observations $(\vx, \vx')$ in a weakly-supervised setting. We sample $\vx$ from the empirical data distribution $p_{\mX}$ and conditionally sample $\vx'$ in an i.i.d. manner, so that $\vx$ and $\vx'$ belong to the same group $g$, i.e.,
\begin{align}
    \vx \; & \sim \; \; p_{\mX}(\vx); \quad \vx' \;  \mathrel{\overset{\makebox[0pt]{\mbox{\normalfont\tiny\sffamily i.i.d.}}}{\sim}}\; \; p(\vx' | \vx, g).
\end{align}

Given $\vx$, we define two latent variables, $\vz_{\Content}$ as content  and $\vz_{\NonContent}$ as style variables. The data
is explained by the generative process:
\begin{align}
    p(\vz_{\Content}) = & \; \mathcal{N}(\vz_{\Content}; 0, \mI); \quad
    p(\vz_{\NonContent}) = \; \mathcal{N}(\vz_{\NonContent}; 0, \mI); \quad
    p(\vx | \vz) = \; p_{\theta}(\vx | \vz_{\Content}, \vz_{\NonContent}) = f_{\theta}(\vx; \vz_{\Content}, \vz_{\NonContent}).
\end{align}
Both $\vz_{\Content}$ and $\vz_{\NonContent}$ are assumed to be independent of each other and are sampled from a Normal distribution with zero mean and diagonal unit variance. $f_{\theta}$ is a suitable likelihood function\footnote{Suitable likelihood functions are, e.g., a Bernoulli likelihood for binary values or a Gaussian likelihood for continuous values.} which is parameterized by a deep neural network. The generative model shown in Figure~\ref{fig:generative} is also known as the decoding part seen in Figure~\ref{fig:autoencoder}.

To perform inference, we approximate the true posterior $p(\vz|\vx)$ with the factorized approximate posterior $q_{\phi}(\vz|\vx)=q_{\phi}(\vz_{\Content}|\vx) \cdot q_{\phi}(\vz_{\NonContent}|\vx)$ that uses a neural network to amortize the the variational parameters. We specify the inference model as
\begin{align}
    q_{\phi}(\vz_{\Content}|\vx) = \mathcal{N}\big(\mu_{\phi,\Content}(\vx), \textrm{diag}(\sigma^2_{\phi,\Content}(\vx))\big); \quad q_{\phi}(\vz_{\NonContent}|\vx) = \mathcal{N}\big(\mu_{\phi,\NonContent}(\vx), \textrm{diag}(\sigma^2_{\phi, \NonContent}(\vx))\big),
\end{align}
where both approximate posteriors are assume to be a factorized Normal distributions with mean $\mu_{\phi}$ and diagonal covariance $\textrm{diag}(\sigma^2_{\phi}(\vx))\big)$.
The inference model is visualized as a graphical model in Figure~\ref{fig:inference} and as the encoding part in Figure~\ref{fig:autoencoder}. The generative and inference models  visualized in Figure~\ref{fig:model_vis} apply to $\vx'$ as well.
\paragraph{VAE objective for paired observation.} Given paired observations $(\vx, \vx')$, the VAE framework maximizes the \pgls{elbo} 
\begin{align}
    \begin{split}
        \textrm{ELBO} = & \; \underbrace{\mathbb{E}_q[\log p(\vx|\vz) + \log p(\vx'|\vz')]}_{\textrm{reconstruction losses}} - \underbrace{\textrm{KL}(q(\vz|\vx) || p(\vz))}_{\substack{\textrm{KL between approximate}\\\textrm{posterior and prior of $\vz$}}} - \underbrace{\textrm{KL}(q(\vz'|\vx') || p(\vz'))}_{\substack{\textrm{KL between approximate}\\\textrm{posterior and prior of $\vz'$}}}\\
        = & \; - \mathcal{L}_{\textrm{pairedVAE}},
    \end{split}\label{eq:vae_paired}
\end{align}
which consists of the reconstruction losses of the observations $\vx$ and $\vx'$ (first two terms) and \pgls{kl} divergence between approximate posterior $q$ and prior $p$ of the latent variables $\vz$ and $\vz'$ (third and fourth term).
This is a straightforward application of the original \pgls{elbo} in \eqref{eq:elbo} to two sets of observations, $\vx$ and $\vx'$.
\paragraph{\pgls{kl} regularization for group similarity.} Rather than defining an average representation for groups as in \citep{bouchacourt2018multi,hosoya2019group}, we propose to enforce consistency between the latent variables by minimizing \pgls{kl} divergence between the latent variables $\vz_{=g}$ and $\vz'_{=g}$. Here, $g$ denotes the group shared between observations $\vx$ and $\vx'$. $\vz_{=g}$ and $\vz'_{=g}$ denote the corresponding group variable, e.g., if $\vx$ and $\vx'$ share group $\Content$ then the corresponding latent variables are $\vz_{\Content}$ and $\vz'_{\Content}$. Given paired observations $\vx, \vx'$ from the same group $g$, our objective is to minimize
\begin{align}
\begin{split}
     \mathcal{L}_{\textrm{KLreg}} = \textrm{KL}\big(q(\vz_{=g}|\vx) || q(\vz'_{=g}|\vx')\big).
\end{split}\label{eq:klreg}
\end{align}
The \pgls{kl} divergence has analytical solutions for Gaussian and Categorical approximate posteriors and is unaffected by the number of shared observations. The analytical solutions can be found in Appendix~\ref{app:groupvae:kl-analytical}.
%
\paragraph{GroupVAE objective and inference.}

\begin{wrapfigure}{R}{0.55\textwidth}
\begin{minipage}{.55\textwidth}
    \vspace{-0.77cm}
    
    \begin{algorithm}[H]
        \caption{GroupVAE Inference}
        \footnotesize{
        \begin{algorithmic}[1]
            \WHILE{ training() }
            \STATE $g^{(1)}, \dots, g^{(n)} \gets \text{getRandomGroups()}$
            \STATE $\mX \gets \text{getMiniBatch}()$
            \STATE $\mX' \gets \text{getPairedObservation(}\mX, g^{(1)}, \dots, g^{(n)}\text{)}$
            {\color{gray} \STATE \# encode $\vx^{(i)}$}
            \STATE $\forall \vx^{(i)} \in \mX: \vz = (\vz^{(i)}_{\Content}, \vz^{(i)}_{\NonContent}) \sim q (\vz^{(i)}_{\Content}, \vz^{(i)}_{\NonContent} | \vx^{(i)})$
            {\color{gray}\STATE \# encode $\vx^{(i)}{'}$}
            \STATE $\forall \vx^{(i)}{'} \in \mX{'}: \vz{'} = (\vz^{(i)}_{\Content}{'}, \vz^{(i)}_{\NonContent}{'}) \sim q (\vz^{(i)}_{\Content}{'}, \vz^{(i)}_{\NonContent}{'} | \vx^{(i)}{'})$
            {\color{gray}\STATE \# calculate loss according to \eqref{eq:groupvae_obj}}
            \STATE $\mathcal{L} \gets \sum_i \mathcal{L}_{\textrm{GroupVAE}}(\vx^{(i)}, \vx'^{(i)},  \vz^{(i)}, \vz'^{(i)}, g^{(i)})  $
            {\color{gray}\STATE \# update gradient $\vg$ and parameters $(\theta, \phi)$}
            \STATE $ (\vg_{\theta}, \vg_{\phi} ) \gets ( \frac{\partial \mathcal{L}}{\partial \theta},  \frac{\partial \mathcal{L}}{\partial \phi} )$
            \STATE $(\theta, \phi) \gets (\theta, \phi) + \alpha (\vg_{\theta}, \vg_{\phi} )$
            \ENDWHILE
        \end{algorithmic}}\label{alg:inference}
    \end{algorithm}
    
\end{minipage}
\end{wrapfigure}
Given a paired observation $(\vx,\vx')$ in the sharing group $g$, we combine the \pgls{elbo} in \eqref{eq:vae_paired} and our proposed \pgls{kl} regularization in \eqref{eq:klreg}. Our proposed model, \textit{GroupVAE}, has the following minimization objective 
\begin{align}
    \mathcal{L}_{\textrm{GroupVAE}} = \mathcal{L}_{\textrm{pairedVAE}} + \gamma \mathcal{L}_{\textrm{KLreg}},\label{eq:groupvae_obj}
\end{align}
where we treat the degree of regularization $\gamma$ as a hyperparameter.
%
We propose an alternating inference strategy to encourage variation in both of the latent variables. If we only utilize observations that belong to one group, e.g., paired observations that always share content, we can obtain a trivial solution for the content latent variable by encoding constant latent variables. We overcome this collapse by alternating the group that the observations belong to during training.
In particular, during inference we randomly sample a group $g \in \{\ucContent, \ucNonContent\}$ and the paired observation $(\vx, \vx')$ according to group g. We then minimize the \pgls{kl} divergence of the corresponding latent variable. The inference's pseudo code is shown in Algorithm~\ref{alg:inference}.

\subsection{Related Work}
\paragraph{Unsupervised learning of disentangled representations.} Various regularization methods for unsupervised disentangled representation learning have been presented in existing works \citep{higgins2016beta,kim2018disentangling,ChenLGD18}. Even though unsupervised methods have shown promising results to learn disentangled representations, \cite{locatello2019challenging} showed in a rigorous study that it is impossible to disentangle factors of variations without any supervision or inductive bias.
Since then, there has been a shift towards weakly-supervised disentanglement learning. Our work follows this stream of works and focuses on the weakly-supervised regime instead of an unsupervised one.

\paragraph{Weakly-supervised learning of disentangled representations.}
\cite{shu2019weakly} investigated different types of weak supervision and provided a theoretical framework to evaluate disentangled representations. \cite{locatello2020weakly} proposed to disentangle groups of variations with only knowing the number of shared groups which can be considered as a complementary component to our method. Similar to our method, both these works follow a weakly-supervised setup. However, both approaches focus on the disentanglement of fine-grained factors, whereas our focus is to disentangle groups.
Before the concept of paired observations was coined by \cite{shu2019weakly} as ``match pairing'', it was already used for geometry and appearance disentanglement \citep{KossaifiTPP18,tran2019disentangling} and group-based disentanglement \citep{bouchacourt2018multi,hosoya2019group}. Closest to our work is MLVAE \citep{bouchacourt2018multi} and GVAE \citep{hosoya2019group}. For group-disentangled representations, MLVAE uses a product of approximate posteriors, whereas GVAE uses an empirical average of the parameters of the approximate posteriors. A thorough analysis of both works is in Appendix~\ref{app:groupvae:gvae_mlvae}. In contrast, we employ a simple and effective KL regularization that has no dependency on the batch size.

\paragraph{Alignment between factors of variations and learned representations.} Closely related to our work and group-based disentanglement concepts are studies that learn specific latent variables corresponding to one or several factors of variations (or labels). \cite{Dupont18} used both continuous and discrete latent variables to improve unsupervised disentanglement of mixed-type latent factors. \cite{creager2019flexibly} proposed to minimize the mutual information between the sensitive latent variable and sensitive labels. Similarly, \cite{KlysSZ18} proposed to minimize \pgls{mi} between the latent variable and a conditional subspace. Both works \citep{creager2019flexibly,KlysSZ18} require either supervision, sensitive labels, or conditions to estimate the mutual information, whereas we only use weak supervision for learning disentangled group representations. Concurrent to our work, \cite{sinha2021consistency} proposed to use a \pgls{kl} regularization for learning a \pgls{vae} with representation that is consistent with augmented data. While \cite{sinha2021consistency} use the KL regularization to enforce the encoding to be consistent with changes in the input, our goal is to split the representation into subspaces that correspond to the different groups of variations.
\section{Evaluation}
Here, we evaluate our GroupVAE and compare it to existing approaches. We show that our approach outperforms existing approaches for group-disentanglement and disentanglement on existing disentanglement benchmarks. Within the context of evaluating group disentanglement, we propose a \pgls{mi}-based evaluation metric to assess the degree of group disentanglement. Further, we demonstrate that our approach is generic and can be applied to various applications, including fair classification and 3D shape-related tasks (reconstruction, classification, and transfer learning).
\subsection{Weakly-supervised group-disentanglement}

\begin{wrapfigure}{}{0.4\textwidth}
\vspace{-0.7cm}
\centering
    \includegraphics[width=0.4\textwidth]{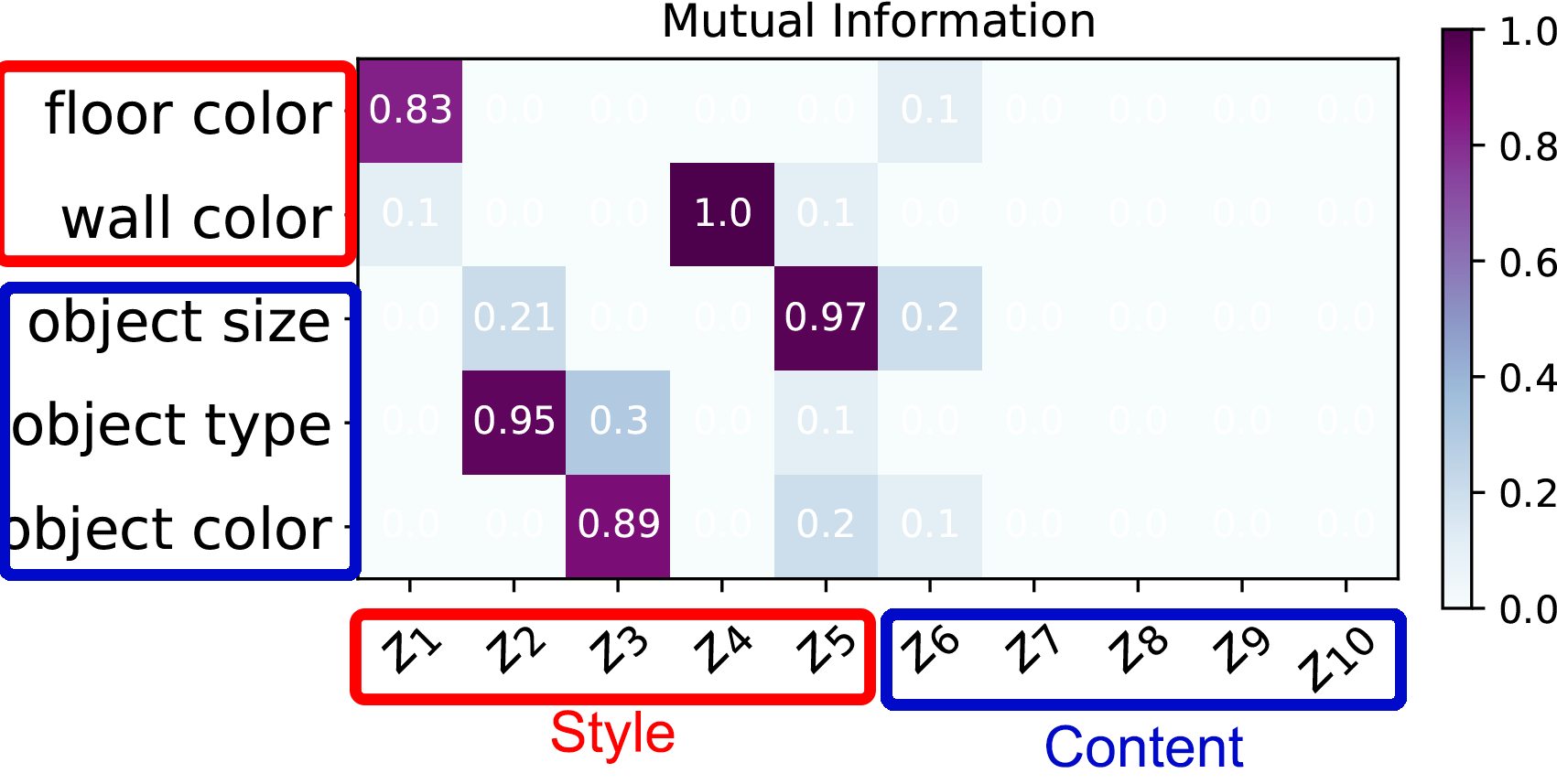}
    \vspace{-0.25cm}
    \caption{\textbf{Example of failed content-style disentanglement with high MIG.} The heatmap shows the MI of each pair of factors and latent dimensions. Although content and style have not been separated in the corresponding latent dimensions, the MIG is still very high ($=0.76$). In contrast, group-MIG considers where the groups are captured, and thus, the group-MIG is much lower ($=0.34$).}
    \label{fig:mig-fail}
\vspace{-0.4cm}
\end{wrapfigure}

\paragraph{Experimental settings.} We used three standard datasets on disentangled representation learning: 3D Cars \citep{reed2014learning}, 3D Shapes \citep{3dshapes18} and dSprites \citep{dsprites17}. Despite the fact that these image datasets are synthetic, disentangling the factors of variation remains a difficult and unresolved task~\citep{locatello2019challenging,locatello2020weakly}. We use \pgls{mig} \citep{ChenLGD18} and our proposed metric \pgls{groupmig} for quantitative evaluation different approaches. We compare our model, GroupVAE, to unsupervised methods ($\beta$-VAE \citep{higgins2016beta} and FactorVAE \citep{kim2018disentangling}) as well as weakly-supervised methods (AdaGVAE \citep{locatello2019challenging}, MLVAE \citep{bouchacourt2018multi}, and GVAE \citep{hosoya2019group}). For all methods, we ran a hyperparameter sweep varying regularization strength for five different seeds. We report the median \pgls{groupmig} and \pgls{mig}. 

\paragraph{\pgls{groupmig} for evaluating group disentanglement.}
The \pacrfull{mig} \citep{ChenLGD18} is a commonly used evaluation metric for disentanglement. 
This metric measures the normalized difference between the latent variable dimensions with highest and second-highest \pgls{mi} values.
The higher the \pgls{mig}, the greater the degree of disentanglement is.
However, \pgls{mig} can still be high if the style latent variable disentangles all factors of variation whereas the content variable collapse to a constant value. An example of a failure in group disentanglement is shown in Figure~\ref{fig:mig-fail}.
Therefore, we introduce \pgls{groupmig}, a metric based on \pgls{mig}, which addresses this issue and quantitatively estimates the mutual information between groups and corresponding latent variables. We define \pgls{groupmig} as
\begin{align}
    \frac{1}{m} \sum_{i=1}^m \frac{1}{H(g_i)} \left | \max I(z_{=g_i}; g_i) - \max_{j \neq i} I(z_{\neq g_j}; g_i) \right |,\label{eq:group-mig}
\end{align}
where $m$ is the number of groups, $g_i$ is the ground truth group, and $I(\vz; g_i)$ is an empirical estimate of the \pgls{mi} between continuous variable $\vz$ and $g_i$. The values of \pgls{groupmig} is small if the group factors are not represented in the corresponding latent vectors, even though the factor is disentangled within the other variables. 
\paragraph{Group labeling.} We define the following groups based on the fine-grained factors for each dataset:
\begin{itemize}
    \item dSprite:s $g_{\ucContent}=[\textrm{shape}, \textrm{scale}], \; g_{\ucNonContent}=[\textrm{orientation}, \textrm{x-position}, \textrm{y-position}]$
    \item 3D Shapes: $g_{\ucContent}=[\textrm{obj. color}, \textrm{obj. size}, \textrm{obj. type}], \; g_{\ucNonContent}=[\textrm{floor color}, \textrm{wall color}, \textrm{azimuth}]$
    \item 3D Cars: $g_{\ucContent}=[\textrm{obj. type}], \; g_{\ucNonContent}=[\textrm{elevation}, \textrm{azimuth}]$
\end{itemize}

\begin{figure}[t]
    \vspace{-0.75cm}
    \centering
    \subcaptionbox{MLVAE}{
        \includegraphics[height=3.3cm]{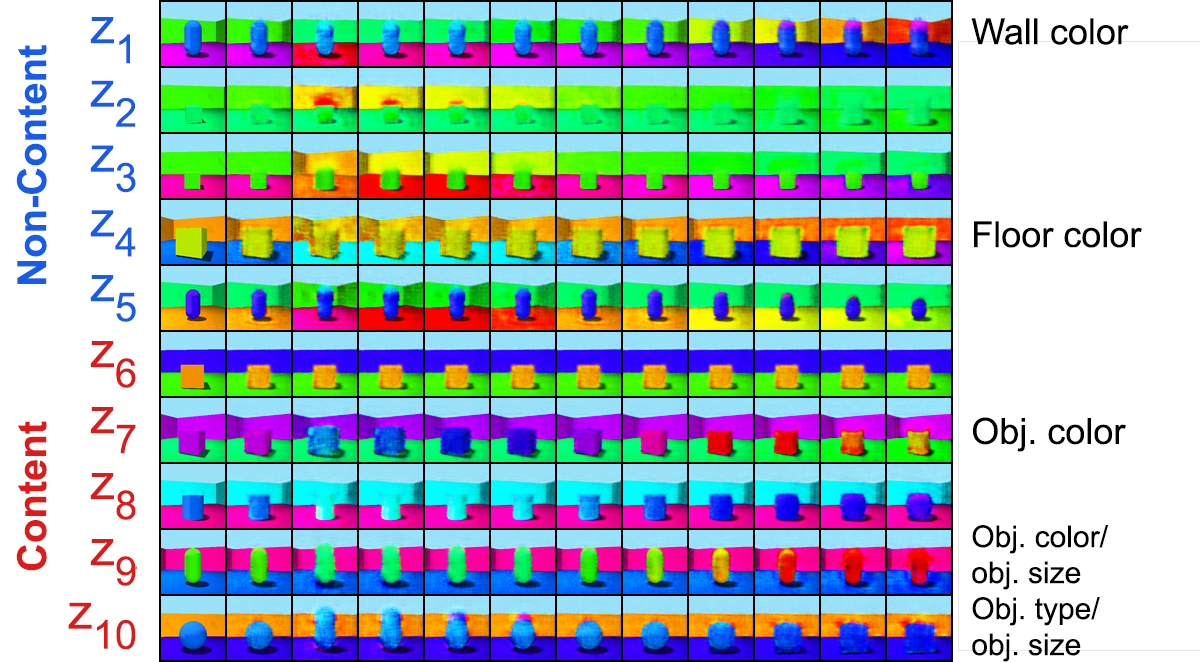}
    }
    \subcaptionbox{GVAE}{
        \includegraphics[height=3.3cm]{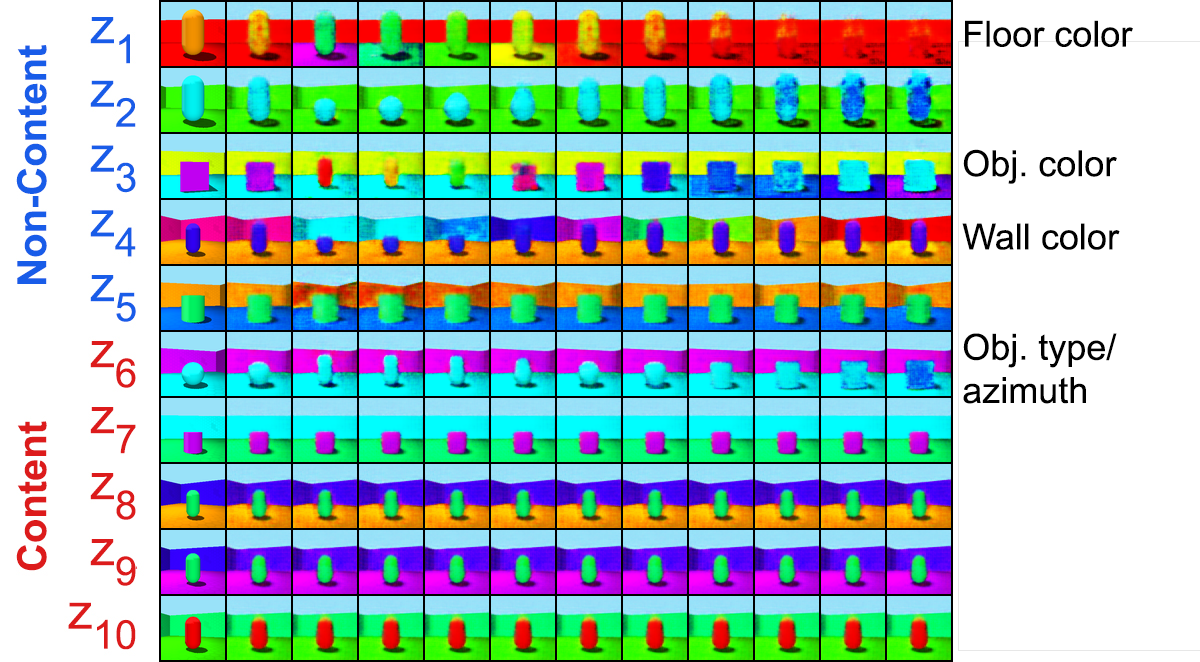}
    }
    \subcaptionbox{GroupVAE (ours)}{
        \includegraphics[height=3.3cm]{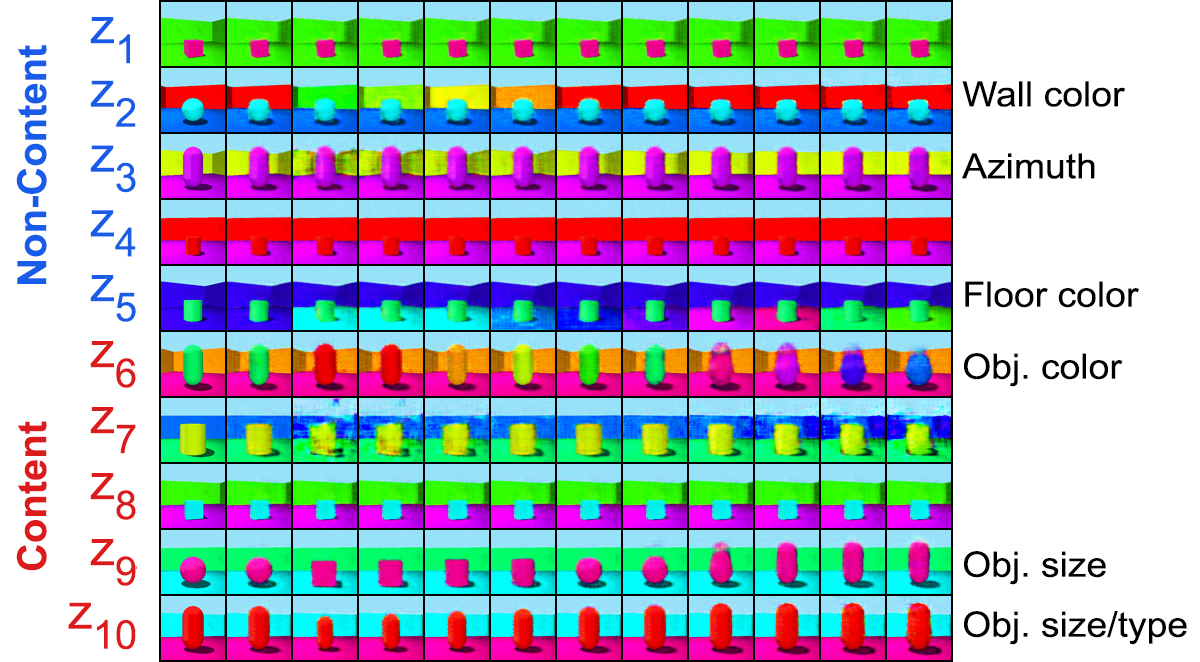}
    }
    \vspace{-0.1cm}
    \caption{\textbf{Interpolations of 3D Shapes.} We show samples from our model GroupVAE and the baseline models (MLVAE and GVAE) with median \pgls{groupmig} over five hyperparameter sweeps. For each subplot, we show random inputs (first column), its reconstructions (second column) and reconstruction when interpolating the latent variables (remaining columns) of each latent dimension (row-wise). The factors annotated on the right side are those with a high level of mutual information ($\textrm{MI} > 0.25$). For all three models $z_1$ to $z_5$ is supposed to capture style (non-content) while $z_5$ to $z_{10}$ is supposed to capture content.}\label{fig:interpolations}
    \vspace{-0.2cm}
\end{figure}

\paragraph{Results.} We consistently outperform weakly-supervised disentanglement models w.r.t. median \pgls{groupmig} over five hyperparameter sweeps of different seeds by \textit{at least 25\%} (3D Shapes). Further, we also improve on disentanglement w.r.t. \pgls{mig} for two out of three datasets (3D Cars, dSprites). 
In addition, we show interpolation samples of MLVAE, GVAE, and GroupVAE\footnote{We selected models with median \pgls{groupmig} over five hyperparameter sweeps of different seeds.} for 3D Shapes in Figure \ref{fig:interpolations}. Both MLVAE and GVAE are not able to capture azimuth in the latent representations. Moreover, GVAE encodes almost all factors into the style part and collapses to a constant representation in the content part. The interpolations of GroupVAE show content and style disentanglement, although some factors such as object size and type for 3D Shapes remain entangled. As we assume that factors in a group can co-variate, this result is expected as object size and type are in the same group. 
%
\begin{table}[t]
    \vspace{-0.5cm}
    \centering
        \caption{\textbf{Quantitative disentanglement results.} We report median \pgls{groupmig} and median \pgls{mig} over five hyperparameter sweeps of different seeds (\textit{higher is better}). Since the unsupervised approaches and AdaGVAE do not learn group disentangled representations, we cannot report \pgls{groupmig} for these groups and denote it with $-$. We highlight in \textbf{bold} the best results.}
    \resizebox{0.8\textwidth}{!}{
        \begin{tabular}{llcccccc}
            \toprule
            & & \multicolumn{2}{c}{3D Cars} & \multicolumn{2}{c}{3D Shapes} & \multicolumn{2}{c}{dSprites}  \\
            Type & Model & \\
            & & group-MIG & MIG & group-MIG & MIG & group-MIG & MIG\\
            \toprule
            unsup. & $\beta$-VAE & $-$ & 0.08 & $-$ & 0.22 & $-$ & 0.10 \\
            unsup. & FactorVAE & $-$ & 0.10 & $-$ & 0.27 & $-$ & 0.14\\
            \midrule
            weakly-sup. & AdaGVAE & $-$ & 0.15 & $-$ & \textbf{0.56} & $-$ & 0.26\\
            weakly-sup. & MLVAE 
                & 0.24 & 0.07
                & 0.47 & 0.32
                & 0.11 & 0.22\\
            weakly-sup. & GVAE 
                &  0.27 & 0.08
                & 0.45 & 0.31 
                & 0.14 & 0.21\\
             \midrule
             weakly-sup. & GroupVAE (ours) 
                & \textbf{0.48} & \textbf{0.18}
                & \textbf{0.60} & 0.31
                & \textbf{0.54} & \textbf{0.27}\\
        \end{tabular}
    }
    \vspace{-0.25cm}
    \label{tab:group-mig}
    
\end{table}
\subsection{Application to fair classification}
\begin{table}[t]
    \vspace{-0.2cm}
    \centering
    \caption{\textbf{Fair classification results on the test set of dSpriteUnfair and CelebA.} We report test accuracy and \pgls{dp} for each sensitive attribute with an average of five experiments. We report the standard error for all test accuracies, but leave out the standard error for all DP results as they were $< 0.002$. We highlight in \textbf{bold} the best results. The column \textit{Fair learning} refers to whether a model uses any supervision during the fair representation learning phase. For the final classification, all models use full supervision.}
    \begin{subtable}{\textwidth}
        \centering
        \resizebox{0.7\textwidth}{!}{
            {
            \scriptsize
            \begin{tabular}{llccc}
                \toprule
                & & & \multicolumn{2}{c}{\textbf{Demographic parity (DP)} $\downarrow \in [0, 1]$}\\
                 \cmidrule{4-5}
                \textbf{Fair learning} & \textbf{Model} & Test acc. $\uparrow$ & ``shape'' $\downarrow$ & ``scale'' $\downarrow$\\
                \midrule
                \ding{55} & MLP &  99.07 {\scriptsize $\pm 0.06$} & 0.007 & 0.008 \\
                \ding{55} & CNN & 99.04 {\scriptsize $\pm 0.05$} & \textbf{0.002} & \textbf{0.002}\\
                \ding{51} (supervised) & FFVAE & 98.60 {\scriptsize $\pm 0.12$} & 0.004 & 0.004 \\
                 \midrule
                 \ding{51} (weakly-superv.) & GroupVAE & \textbf{99.18} {\scriptsize $\pm 0.08$} & \textbf{0.002} & \textbf{0.002}\\
            \end{tabular}
            }
        }
        \caption{Results for dSpritesUnfair predicting ``x-position''.}
    \end{subtable}
    \begin{subtable}{\textwidth}
        \centering
        \resizebox{0.7\textwidth}{!}{
            {
            \scriptsize
            \begin{tabular}{llccc}
                \toprule
                & & & \multicolumn{2}{c}{\textbf{Demographic parity (DP)} $\downarrow \in [0, 1]$}\\
                \cmidrule{4-5}
                \textbf{Fair learning} & \textbf{Model} & \textbf{Test acc.} $\uparrow$  & ``Male'' & ``Young'' \\
                \midrule
                \ding{55} & MLP & 97.89 {\scriptsize $\pm 0.01$} & 0.99 & 0.99 \\
                \ding{55} & CNN & \textbf{98.46} {\scriptsize $\pm 0.03$} &  0.95  & 0.93\\
                \ding{51} (supervised) & FFVAE & 97.79 {\scriptsize $\pm 0.01$}& 0.04 & 0.04 \\
                 \midrule
                 \ding{51} (weakly-superv.) & GroupVAE & 98.23 {\scriptsize $\pm 0.02$} & \textbf{0.01} & \textbf{0.02}\\
            \end{tabular}
            }
        }
        \caption{Results for CelebA predicting ``bald''.}
    \end{subtable}
    \begin{subtable}{0.95\textwidth}
        \resizebox{\textwidth}{!}{
            {
            \scriptsize
            \begin{tabular}{llcccccc}
                \toprule
                & & & \multicolumn{4}{c}{\textbf{Demographic parity (DP)} $\downarrow \in [0, 1]$}\\
                \cmidrule{4-7}
                \textbf{Fair learning} & \textbf{Model} & \textbf{Test acc.} $\uparrow$ & `BigNose'  & `HeavyMakeup' & `Male' & `WearingLipstick'\\
                \midrule
                \ding{55} & MLP & 77.24 {\scriptsize $\pm 0.29$} & 0.09 & 0.15 & 0.06 & 0.04 \\
                \ding{55} & CNN & 79.90 {\scriptsize $\pm 0.06$} &  0.11 & 0.15 & 0.03 & 0.06\\
                \ding{51} (supervised) & FFVAE & 97.75 {\scriptsize $\pm 0.03$}& 0.03 & \textbf{0.02} & 0.03 & 0.03\\
                 \midrule
                 \ding{51} (weakly-superv.) & GroupVAE & \textbf{97.88} {\scriptsize $\pm 0.01$} & \textbf{0.01} & \textbf{0.02} & \textbf{0.02} & \textbf{0.01}\\
            \end{tabular}
            }
        }
        \caption{Results for CelebA predicting ``attractive''.}
        \end{subtable}%
    \vspace{-0.25cm}
    \label{table:fair}
\end{table}
We examine the problem of learning fair representations for classification problems as an application of our method. In particular, we want to learn fair group representation in which members of any (demographic) groups have an equal probability of being assigned to the positive predicted class.
Deep learning algorithms have been proven to be biased against specific demographic groups or populations \citep{mehrabi2021survey}. It is critical that classification models can produce accurate predictions without discriminating against certain groups in high-stakes and safe-related applications. In this context, we propose to learn fair representations by learning two distinct groups of representations: a predictive representation for evaluating the downstream task and a representation to account for the sensitive factors, e.g., gender- or age-specific attributes. The latter representation is solely utilized for training and not for downstream tasks.

Learning fair representations consist of a two-step optimization scheme. First, we train GroupVAE with pairs of observations sharing either sensitive and non-sensitive attributes. Second, we train a simple MLP for attribute classification using the non-sensitive mean representation. We measure classification accuracy and \pacrfull{dp}. \pgls{dp} measures whether the predictive outcome is independent of a sensitive attribute. A completely fair model would attain a \pgls{dp} value of 0.0, whereas a biased model can have a \pgls{dp} up to 1.0. We compare against MLP and CNN baselines, and FFVAE \pcite{creager2019flexibly} which learns fair representations by using a supervised loss on the sensitive attributes and a total correlation loss.
We used two datasets: dSpritesUnfair \pcite{creager2019flexibly,trauble2020independence} and CelebA \pcite{liu2015deep}. dSpritesUnfair is a modified image dataset based on dSprites with binarized factors of variations and is sampled with shape and x-position being highly correlated.
For CelebA, an image dataset of celebrity faces with 40 binary attribute labels, 
we predict ``bald'' and ``attractive'' in two separate experiments. For predicting ``bald'', we use the attributes ``male" and ``young"  as sensitive attributes whereas we use the attributes ``BigNose'', ``HeavyMakeUp'', ``Male'' and ``WearingLipstick'' as sensitive attributes for predicting ``attractive''. We argue that these attributes have a weak correlation but a strong correlation with the predictive attribute. However, several CelebA attributes significantly correlate, making this a difficult dataset for fairness classification.
We refer to the Appendix~\ref{app:groupvae:fair-exp-setting} for the detailed experimental settings.

\paragraph{Results.}
We report the fair classification results in Table~\ref{table:fair}. Overall, the results in Table~\ref{table:fair} show that weakly-supervised fair representation learning (GroupVAE) outperforms supervised fair representation learning (FFVAE). Further, we either get competitive or even outperform the supervised baselines (MLP, CNN).
Surprisingly, when evaluating dSpritesUnfair the demographic parity for all models is relatively low, and the strong correlation between shape and x-position does not seem to affect the classification. The test accuracy and \pgls{dp} of the sensitive attributes of all the competitive models are very close to each other. Nevertheless, among all models, our method achieves the highest test accuracy and lowest \pglspl{dp}. For predicting ``bald'' in CelebA, even though both MLP and CNN baselines achieve high test accuracy, the \pglspl{dp} shows an extremely biased classification towards gender-specific and male-specific attributes. In contrast, our GroupVAE achieves the lowest \pglspl{dp} but still attain competitive classification accuracy, i.e., second highest test accuracy after the CNN performance. When predicting ``attractive'', GroupVAE decreases the bias of all sensitive attributes and increases the test accuracy compared to all other models.
\subsection{Application to 3D point cloud tasks}

In addition to evaluating image datasets, we show experiments on 3D point clouds for reconstruction and classification. We experimented with FoldingNet~\pcite{yang2018foldingnet}, a deep autoencoder that learns to reconstruct 3D point clouds in an unsupervised way. Unlike VAEs, the FoldingNet autoencoder is deterministic and does not optimize the representation to be a probabilistic distribution. Instead of converting the autoencoder into a VAE, we use a similar approach as \cite{ghosh2019variational}. We assume the embedding of autoencoder to be Normal distributed with constant variance. Given this assumption, the KL divergence between the corresponding embeddings reduces to a simple L2 regularization, and we can inject noise to regularize the decoding. We evaluate three tasks, 3D point cloud reconstruction, classification, and transfer learning. We measure the Chamfer Distance (CD) and the Earth Mover's Distance (EMD) to assess reconstruction quality and report accuracy to assess classification and transfer learning performance. We compare to FoldingNet (unsupervised) and DGCNN (supervised) \pcite{wang2019dynamic}, a dynamic graph-based classification approach. For assessing the transfer learning capability, we use a linear SVM classifier on the extracted representation. We used two datasets for training: FG3D \pcite{liu2021fine} and ShapeNetV2 \pcite{chang2015shapenet}. FG3D contains 24{,}730 shapes with annotations of basic categories (Airplane, Car, and Chair) and fine-grained sub-categories. ShapeNetV2 contains 51{,}127 shapes with annotations of 55 categories. For transfer learning, we also use ModelNet40 \pcite{wu20153d}.

\begin{figure}
    \vspace{-0.5cm}
    \centering
    \subcaptionbox{Reconstructions.}{
    \includegraphics[width=0.35\textwidth]{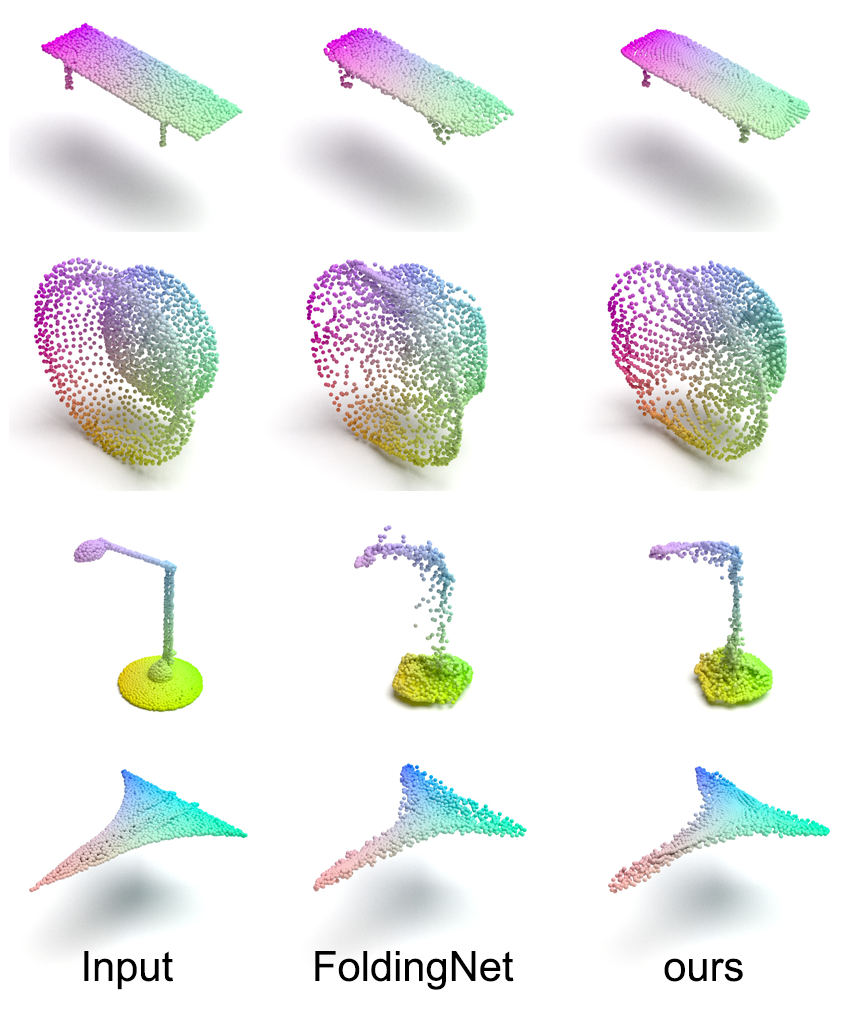}
    }
    \subcaptionbox{Interpolations between two different samples.}{
     \includegraphics[width=0.6\textwidth]{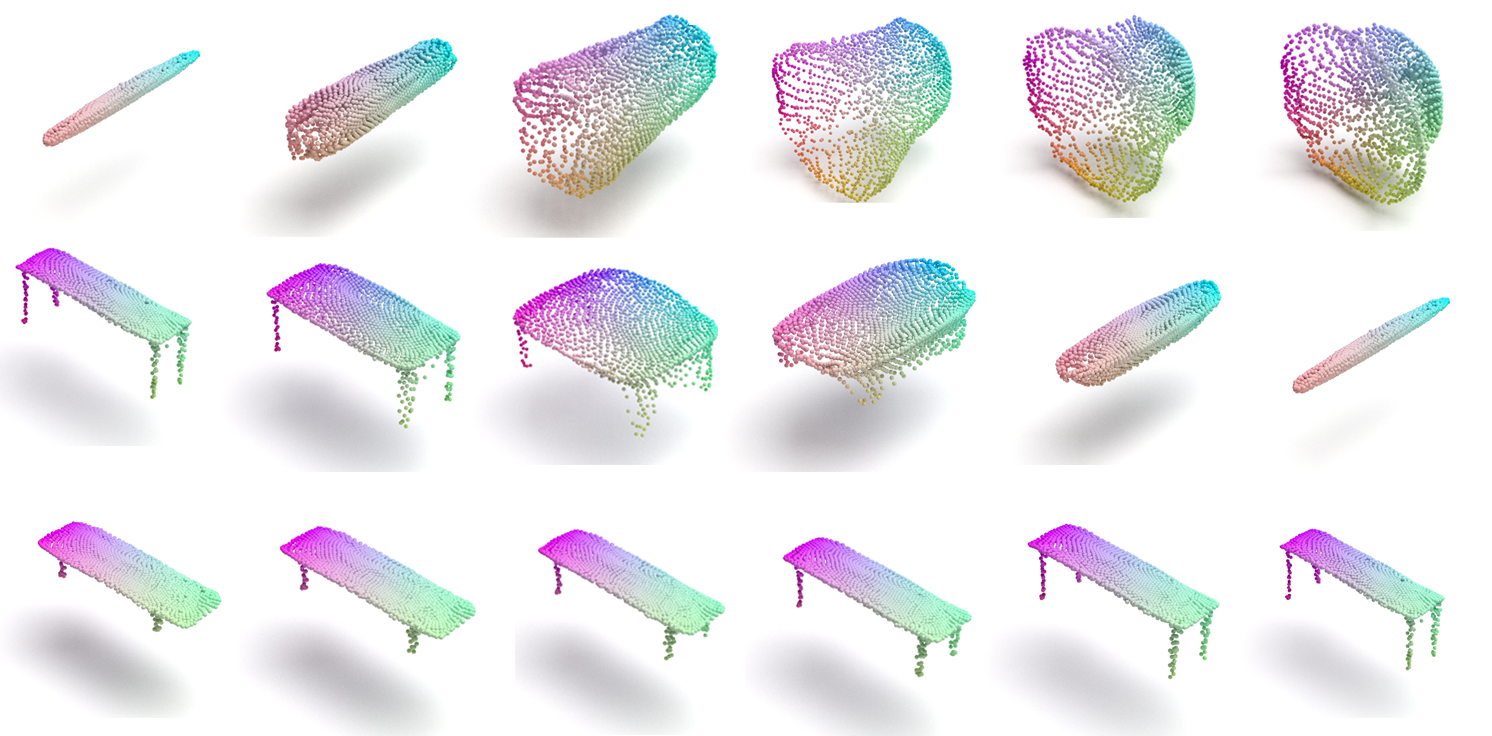}
     \vspace{0.75cm}
    }
    \caption{\textbf{Qualitative samples of ShapeNetV2.} We show reconstructions of FoldingNet and our approach in (a) and show interpolations of our approach in (b).} 
    \label{fig:pc}
    \vspace{-0.25cm}
\end{figure}

\paragraph{Results.} Table~\ref{tab:3dvae:recon} shows that weakly-supervised training improves upon 3D point cloud reconstruction for both FG3D and ShapeNetV2.
Table~\ref{tab:3dvae:transfer} shows the classification and transfer results. Our approach GroupFoldingNet improves point cloud classification compared to the original FoldingNet and is competitive with the supervised approach when training with FG3D.
We outperform both supervised and unsupervised transfer learning performances when training with FG3D and evaluating ShapeNetV2 and ModelNet40. We are competitive to the supervised approach when training with ShapeNetV2 and evaluating on ModelNet40.
In particular, the transfer learning performance with FG3D as the training set highlights the capabilities of weakly-supervised group disentanglement as it can learn 3D point clouds of three classes and transfer it to ShapeNetV2, a large-scale dataset with 55 classes.
We also visualize point cloud reconstructions and interpolations of three different classes using our approach in Figure~\ref{fig:pc}. The reconstructions show that our approach is better than FoldingNet in reconstructing finer details. Further, the interpolations show that our approach can learn an interpretable representation.

\begin{table}[t]
    \vspace{-0.5cm}
    \centering
    \caption{\textbf{Evaluation of 3D point cloud reconstruction, classification, and transfer learning.} We report Chamfer Distance (CD) and Earth Mover Distance (EMD) for quality of reconstruction and accuracy for classification and transfer learning tasks. Best results without full supervision are highlighted in \textbf{bold}.}
    \begin{subtable}{\textwidth}
    \centering
    \resizebox{0.75\textwidth}{!}{
    {\scriptsize
        \begin{tabular}{llcccccc}
            \toprule
            & & \multicolumn{5}{c}{\textbf{Reconstruction $\downarrow$}} \\
            \cmidrule{3-7}
            & & \multicolumn{2}{c}{FG3D} & & \multicolumn{2}{c}{ShapeNetV2}\\
            \textbf{Type} & \textbf{Model} & CD & EMD & & CD & EMD \\
            \midrule
            unsupervised & FoldingNet & 0.9539 & 0.9340 & & 2.9867 & 1.5576\\
            weakly-superv. & GroupFoldingNet (ours) & \textbf{0.7519} & \textbf{0.8191} & & \textbf{2.6891} & \textbf{1.3009}\\
        \end{tabular}
    }
    }
    \caption{Reconstruction results for FG3D and ShapeNetV2.}\label{tab:3dvae:recon}
    \end{subtable}
    \begin{subtable}{\textwidth}
    \centering
    \resizebox{\textwidth}{!}{
    {\scriptsize
        \begin{tabular}{llccccccc}
            \toprule
            & & & & & & &\multicolumn{2}{c}{\textbf{Linear SVM ACC $\uparrow$}}\\
            \cmidrule{8-9}
            \textbf{Type} & \textbf{Model} & \textbf{Training dataset} & \textbf{Test dataset} & \textbf{\#classes} & \textbf{Test ACC} $\uparrow$ & &  ShapeNetV2 & ModelNet40 \\
            \midrule
            supervised & DGCNN & FG3D & FG3D & 3 & 99.26 & & 50.53 & 74.25\\
            \hdashline
            unsupervised & FoldingNet & FG3D & FG3D & 3 & 98.27 & & 85.45 & 80.04\\
            weakly-superv. & ours & FG3D & FG3D & 3 & \textbf{98.57} && \textbf{87.24} & \textbf{81.39}\\
            \midrule
            supervised & DGCNN & ShapeNetV2 & ShapeNetV2 & 55 & 94.4 & & $-$ & 90.02\\
            \hdashline
            unsupervised & FoldingNet & ShapeNetV2 & ShapeNetV2 & 55 & 81.51 & & $-$ & 87.40\\
            weakly-superv. & ours & ShapeNetV2 & ShapeNetV2 & 55 & \textbf{82.62} & & $-$ & \textbf{89.97}\\
        \end{tabular}
    }
    }
    \caption{Classification and transfer learning of representations.}\label{tab:3dvae:transfer}
    \end{subtable}
    \label{tab:3dvae}
\end{table}

\section{Conclusion \& discussion}
We proposed a simple KL regularization for VAEs to enforce group disentanglement through weak supervision. We empirically showed that our model outperforms existing approaches in group disentanglement. Further, we demonstrated that learning group-disentangled representations outperforms performance on fair image classification and 3D shape-related tasks (reconstruction, classification, and transfer learning) and is even competitive to supervised approaches.

There are several possible directions for future work.
In comparison to unsupervised representation learning, weakly-supervised learning, by definition, requires some weak form of supervision. Although we only need knowledge of whether two observations share a specific group, this limits the approach. Further, we require group labels for the entire dataset for training and evaluation. For real-life applications, datasets may not be fully labeled, and performance may suffer under this setting. Future investigation of group disentanglement in a low data or a ``semi'' weakly-supervised regime can allow group disentanglement learning to transfer to large-scale and more realistic settings.
Another promising direction is investigating models with more than two groups. Even though we chose to focus on applications with two groups in this work, our method can generalize to more than two groups, which is a promising direction for future work.


\

\subsubsection*{Acknowledgments}
We thank Hooman Shayani and Tonya Custis for useful discussions and comments on the paper.

\bibliography{bib/groupvae}
\bibliographystyle{sty/iclr2022_conference}
\clearpage

\appendix
\section{GroupVAE}\label{app:groupvae:groupvae-deriv}

\subsection{Joint Learning of Continuous and Discrete Groups}\label{app:groupvae:joint}

The generative model defined in the main Section 4 assumes both content and style representations to be Gaussian distributed. However, many data-generating processes rely on discrete factors which is usually difficult to capture with continuous variables. In these cases, we can define the generative model as
\begin{align}
    p(\vz_{\Content}) = & \; \textrm{Cat}(\pi),\\
    p(\vz_{\NonContent}) = & \; \mathcal{N}(0, \mI),\\
    p(\vx | \vz_{\Content}, \vz_{\NonContent}) = & \; \textrm{Bernoulli}(f_{\theta}(\vz_{\Content}, \vz_{\NonContent})).
\end{align}
For inference, we use a Gumbel-Softmax reparameterization \cite{JangGP17,MaddisonMT17}, a continuous distribution on the simplex that can approximate categorical samples for $\vz_{\Content}$. Similar to the \gls{kl} divergence between two Normal distributions, the \gls{kl} divergence between two Categorical distributions can also be computed in closed form. 
\subsection{Closed-form Solutions for the KL Regularization}\label{app:groupvae:kl-analytical}
In the case of both $\vz \sim \mathcal N(\mu, \sigma^2)$ and  $\vz' \sim \mathcal N(\mu', \sigma'^2)$ being factorized Gaussian distributions, the KL regularization has the analytical solution 
\begin{align}
    \textrm{KL}(\vz||\vz') = \log \frac{\sigma'}{\sigma} + \frac{\sigma^2 + (\mu - \mu')^2}{2\sigma'^2} - \frac{1}{2}.
\end{align}
In the case of $\vz ~ \textrm{Categorical}(\pi)$ and $\vz' ~ \textrm{Categorical}(\pi')$, the KL has the analytical solution
\begin{align}
    \textrm{KL}(\vz||\vz') = \sum_i \pi_i \log \frac{\pi_i}{\pi'_i}.
\end{align}

\section{Analysis of Existing Group-Disentanglement Approaches}\label{app:groupvae:gvae_mlvae}
In this Section, we give further details about the content approximate posterior proposed by Bouchacourt et al. \cite{bouchacourt2018multi}, and Hosoya \cite{hosoya2019group}. Further, we analyze the proposed approaches and show its limitations.

\subsection{MLVAE and GVAE}
As described in Subsection~\ref{sec:groupvae}, we restrict to two groups and define corresponding latent variables $\vz_{\Content}$ and $\vz_{\NonContent}$ given observation $\vx$\footnote{In similar fashion, we define two latent variables $\vz'_{\Content}$ and $\vz'_{\NonContent}$ for observation $\vx'$.}. However, both works also apply to any number of groups.
For paired observations $(\vx, \vx')$ with shared group factor $c$, the loss objectives for MLVAE \pcite{bouchacourt2018multi} and GVAE \pcite{hosoya2019group} are
\begin{align}
    \mathcal{L}_{\textrm{MLVAE}} = & \mathcal{L}_{\textrm{pairedVAE}} - \beta \textrm{KL}(q_{\phi}(\tilde{\vz}_{c}, \vz_{ \NonContent} | \vx) || p(\vz)) - \beta \textrm{KL}(q_{\phi}(\tilde{\vz}_{c}, \vz_{ \NonContent'}| \vx') || p(\vz)), \label{eq:mlvae}\\
    \mathcal{L}_{\textrm{GVAE}} = & \mathcal{L}_{\textrm{pairedVAE}} - \beta \textrm{KL}(q_{\phi}(\bar{\vz}_{c}, \vz_{ \NonContent} | \vx) || p(\vz)) - \beta \textrm{KL}(q_{\phi}(\bar{\vz}_{c}, \vz_{ \NonContent'}|\vx') || p(\vz)).\label{eq:gvae}
\end{align}
The loss objectives $\mathcal{L}_{\textrm{GVAE}}$ and $\mathcal{L}_{\textrm{MLVAE}}$ are very similar. The only exceptions are the group approximate posteriors, $\bar{\vz}_c$ for $\mathcal{L}_{\textrm{GVAE}}$ and $\tilde{\vz}_c$ for $\mathcal{L}_{\textrm{GVAE}}$.

\cite{bouchacourt2018multi} assume the group approximate posterior to be a product of the individual approximate posteriors sharing the same group $\bar{\vz}_{\Content}$
\begin{align}
    \bar{\vz}_{\Content} \sim \mathcal{N}(\mu_{\phi,\Content}(\vx), \textrm{diag}(\sigma^2_{\phi,\Content}(\vx)) \cdot \mathcal{N}(\mu_{\phi,\Content}(\vx'), \textrm{diag}(\sigma_{\phi,\Content}^2(\vx'))).
\end{align}

The product of two or more Normal distributions is Normal distributed, and thus the \pgls{kl} term can still be calculated in closed-form.

\cite{hosoya2019group} uses an empirical average over the parameters of the individual approximate posteriors. The group approximate posterior is defined as
\begin{align}
    \tilde{z}_{\Content} \sim \mathcal{N}(0.5 \cdot \mu_{\phi,\Content}(\vx) + 0.5 \cdot \mu_{\phi,\Content}(\vx'), 0.5 \cdot \textrm{diag}(\sigma_{\phi,\Content}^2(\vx)) + 0.5 \cdot \textrm{diag}(\sigma_{\phi,\Content}^2(\vx'))).
\end{align}
\subsection{Analysis}
Both MLVAE and GVAE enforce disentanglement through the $\beta$-regularization in the last two terms of \eqrefpl{eq:mlvae}{eq:gvae}. This regularization was also used in $\beta$-VAE \cite{higgins2016beta} which regularizes a trade-off between disentanglement and reconstruction. The two \pgls{kl} terms in \eqrefpl{eq:mlvae}{eq:gvae} can be decomposed similar to the \pgls{elbo} and \pgls{kl} decomposition in \cite{hoffman2016elbo,ChenLGD18}. We consider the objective in \eqref{eq:mlvae} averaged over the empirical distribution $p(n)$. Each training sample denoted by a unique index and treated as random variable $n$. We simplify $q(\vz|\vx_n) = q(\vz|n)$ and refer to $q(\vz) = \sum_i^N q(\vz|n)p(n)$ as the aggregated posterior \cite{hoffman2016elbo}. We can decompose the first \pgls{kl} in \eqref{eq:mlvae}\footnote{We can decompose the \pgls{kl} of GVAE in \eqref{eq:gvae} similarly.} as 
\begin{align}
    \begin{split}
         &\; \E_{p(n)} \bigg[ \textrm{KL}\big(q(\bar{\vz}_c, \vz_{\NonContent}|n) || p(\vz)\big)\bigg] \\
        = &\; \textrm{KL}\big( q(\bar{\vz}_{\Content}, \vz_{\NonContent}, n) || q(\bar{\vz}_{\Content}, \vz_{\NonContent}) p(n) \big) + \sum_j \textrm{KL}\big(q(\bar{\vz}_j) || p(\vz_j) \big) \\
        & \; + \underbrace{\textrm{KL} \big( q(\bar{\vz}_{\Content}) || \prod_j q(\vz_{\Content,j})\big)}_{\textrm{content total correlation}}  + \underbrace{\textrm{KL} \big( q(\vz_{\NonContent} ||  \prod_j q(\vz_{\NonContent,j})) \big)}_{\textrm{style total correlation}},
    \end{split}\label{eq:mlvae-decomp-kl}
\end{align}
where $q(\bar{\vz}) = q(\bar{\vz}_{\Content}, \vz_{\NonContent}) = q(\bar{\vz}_{\Content})\cdot q(\vz_{\NonContent})$. We show the full derivation in the next Subsection~\ref{app:groupvae:mlvae-gvae-kl-decomp}. Minimizing the averaged \pgls{kl} between the content and style latent variables ($q(\bar{\vz}_c, \vz_{\NonContent}|n)$) and the prior $p(\vz)$ also leads to minimization of the total correlation of content variables and style variables (the last two terms in \eqref{eq:mlvae-decomp-kl}). The total correlation quantifies the amount of information shared between multiple random variables, i.e., low total correlation indicates high independence between the variables. 
Even though this objective motivates disentangled content and style representations, the group representation depends on the number of samples used for the averaging.
Further, both \cite{bouchacourt2018multi} and \cite{hosoya2019group} only average over the content group. There are no structural nor optimization constraints that prevent the style latent variable from encoding all factors of variation.
%
\begin{figure*}[t]
    \centering
    \subcaptionbox{3DShapes: MI between latent dimensions and factors of variation of a trained GVAE model with $\textrm{MIG}=0.55$ and $\textrm{group-MIG}=0.44$.}{
        \vspace{0.1cm}
        \includegraphics[width=0.41\textwidth]{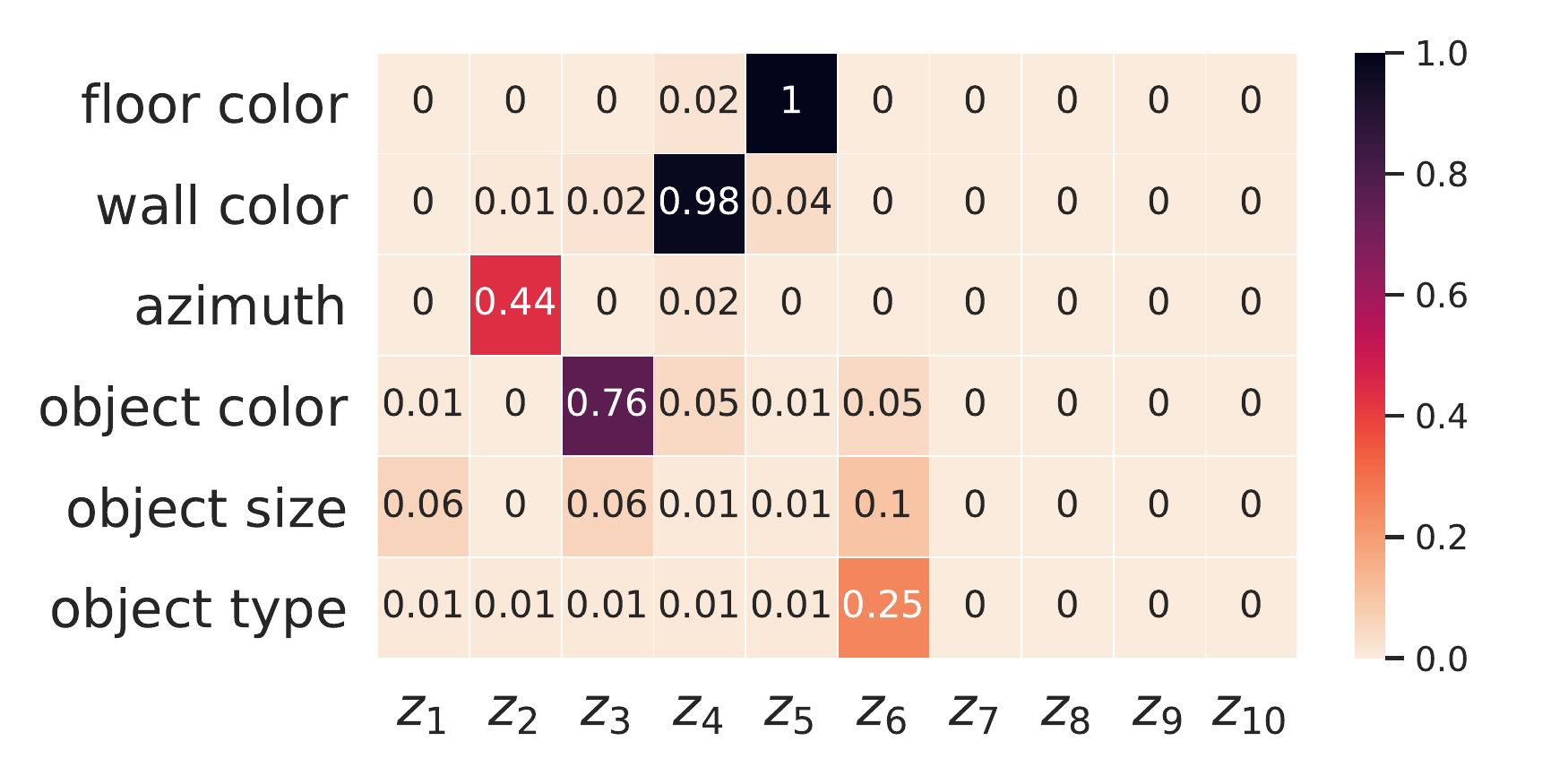} 
    }
    \hspace{0.1cm}
    \subcaptionbox{dSprites: group-MIG of content and style information for all hyperparameter runs.}{
        \includegraphics[width=0.26\textwidth]{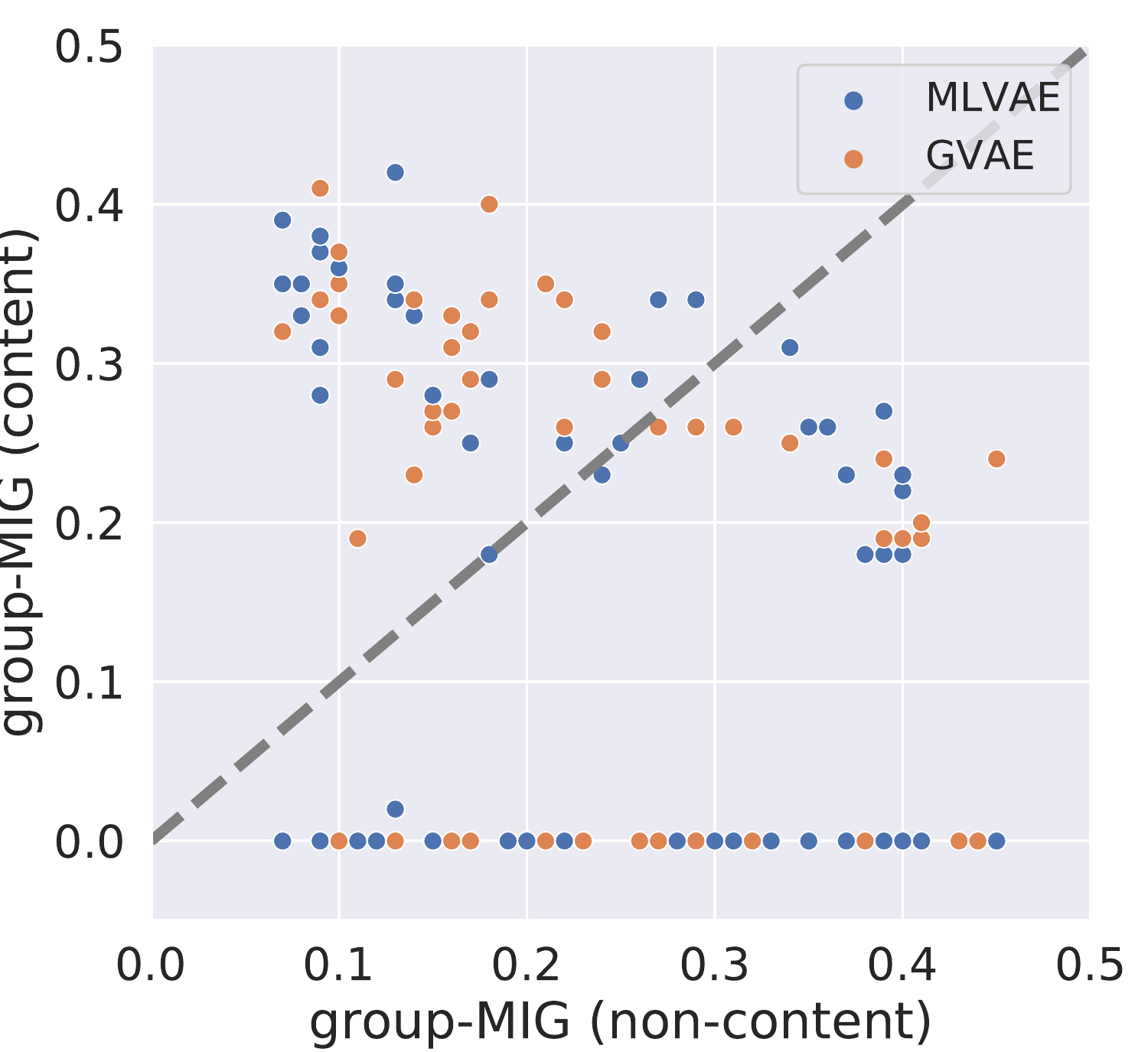}  
    }
    \hspace{0.1cm}
    \subcaptionbox{
        dSprites: MIG w.r.t. different number of shared observations for MLVAE and GVAE. 
    }{\includegraphics[width=0.26\textwidth]{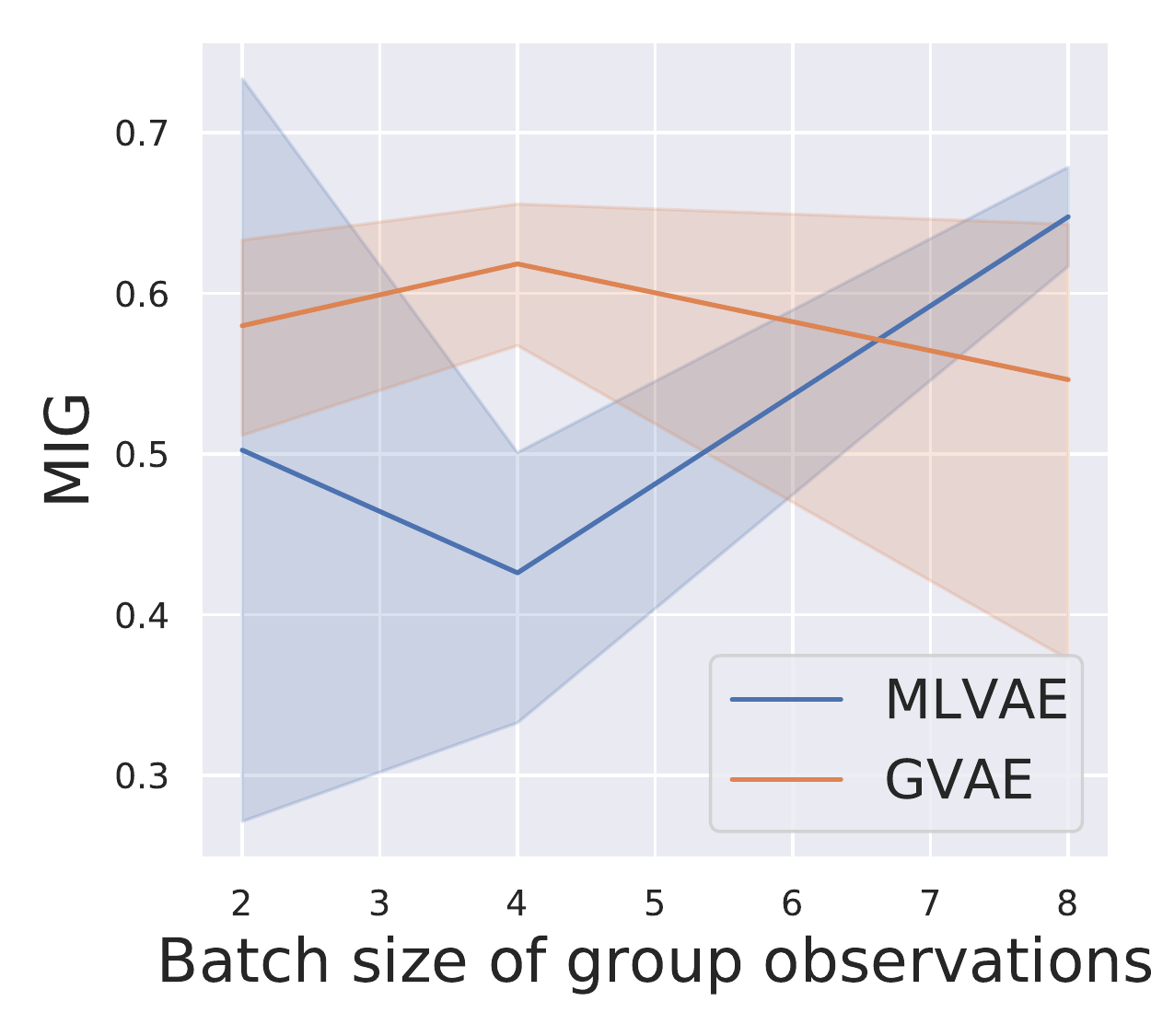}}
\caption{\textbf{Collapse and sensitivity of existing weakly supervised group disentanglement models.} (a) shows mutual information (MI, \textit{higher is better}) for a GVAE model trained on 3DShapes. (b) plots both group-MIG (\textit{higher is better}) w.r.t. content and style information trained on dSprites. (c) plots MIG (\textit{higher is better}) w.r.t. number of shared observations.}\label{fig:weak-collapse}
\end{figure*}
\paragraph{Sensitivity to group batch size.} MLVAE and GVAE use different types of averaging over group latent variables. In realistic settings, always having a certain number of observations that share the same group variations can be difficult. For instance, when training MLVAE and GVAE with dSprites, the performance and its variance is correlated with the number of shared observations. We visualized these findings in Figure \ref{fig:weak-collapse} (c).
\paragraph{Visualization of collapse.} We visualize such behavior in Figure \ref{fig:weak-collapse} (a) on a GVAE model trained on 3DShapes with two groups of variations $\Content=$\{object color, object size and object type\} and $\NonContent=$ \{floor color, wall color, azimuth\}. Ideally, $\vz_1-\vz_5$ contains high mutual information with group factors $\NonContent$ and $\vz_6-\vz_{10}$ contains high mutual information with group factors $\Content$. However, most information is captured in $\vz_1-\vz_5$, whereas only a little information about object type is contained in $\vz_6-\vz_{10}$.

\subsection{KL Decomposition}\label{app:groupvae:mlvae-gvae-kl-decomp}
Here, we show the full derivation for ~\eqref{eq:mlvae-decomp-kl}. For a given group $g$ the KL decompose as follows:
\begin{align}
    & \; \E_{p(n)} \bigg[ \textrm{KL}\big(q(\bar{\vz}_c, \vz_{\NonContent}|n) || p(\vz)\big)\bigg] \\
    = & \; \mathbb{E}_{p(n)}\bigg[\mathbb{E}_{q(\bar{\vz}_{\Content}, \vz_{\NonContent}|n)}\big[\log q(\bar{\vz}_{\Content}, \vz_{\NonContent}|n) - \log p(\vz) \big]\bigg]\\
    \begin{split}
         = & \; \mathbb{E}_{p(n)}\bigg[\mathbb{E}_{q(\bar{\vz}_{\Content}, \vz_{\NonContent}|n)}\big[\log q(\bar{\vz}_{\Content}, \vz_{\NonContent}|n) - \log p(\vz)
         + \underbrace{\log q(\bar{\vz}_{\Content}, \vz_{\NonContent}) - \log q(\bar{\vz}_{\Content}, \vz_{\NonContent})}_{=0} \\
         & \; + \underbrace{\log \prod_{i} q(\bar{\vz_i}) - \log [\prod_{i} q(\bar{\vz}_{\Content,i}) \prod_{j} q(\vz_{\NonContent,j}) }_{=0} \big]\bigg]
    \end{split}\\
    \begin{split}
    = &\; \mathbb{E}_{q(\bar{\vz}_{\Content}, \vz_{\NonContent}, n)} \bigg[ \log \frac{q(\bar{\vz_{\Content}}, \vz_{\NonContent}|n)}{q(\bar{\vz}_{\Content}, \vz_{\NonContent})} \bigg] + \mathbb{E}_{q(\bar{\vz}_{\Content}, \vz_{\NonContent})}\bigg[\log \frac{q(\bar{\vz}_{\Content}, \vz_{\NonContent})}{\prod_{i} q(\bar{\vz}_{\Content,i}) \prod_{j} q(\vz_{\NonContent,j})}\bigg]\\
     & \; + \mathbb{E}_{q(\bar{\vz})}\bigg[\log \frac{\prod_i q(\bar{\vz}_i)}{p(\vz)} \bigg]
    \end{split}\\
    \begin{split}
    = &\; \mathbb{E}_{q(\bar{\vz}_{\Content}, \vz_{\NonContent}, n)} \bigg[ \log \frac{q(\bar{\vz_{\Content}}, \vz_{\NonContent}|n)p(n)}{q(\bar{\vz}_{\Content}, \vz_{\NonContent})p(n)} \bigg]
    + \mathbb{E}_{q(\bar{\vz}_{\Content})}\bigg[\log \frac{q(\bar{\vz}_{\Content})}{\prod_{i} q(\bar{\vz}_{\Content,i})}\bigg] \\
    & \; + \mathbb{E}_{q(\vz_{\NonContent})}\bigg[\log \frac{q(\bar{\vz}_{\NonContent})}{\prod_{j} q(\vz_{\NonContent,j})}\bigg] + \mathbb{E}_{q(\bar{\vz})}\bigg[\sum_i \log \frac{q(\bar{\vz}_i)}{p(\vz_i)} \bigg]
    \end{split}\\
    \begin{split}
    = &\; \underbrace{\textrm{KL}\big( q(\bar{\vz}_{\Content}, \vz_{\NonContent}, n) || q(\bar{\vz}_{\Content}, \vz_{\NonContent}) p(n) \big)}_{\textrm{Index-code MI}} + \underbrace{\textrm{KL} \big( q(\bar{\vz}_{\Content}) || \prod_i q(\vz_{\Content,i})\big)}_{\textrm{content total correlation}}  + \underbrace{\textrm{KL} \big( q(\vz_{\NonContent} ||  \prod_i q(\vz_{\NonContent,i})) \big)}_{\textrm{style total correlation}}\\
    & \; + \underbrace{\sum_i \textrm{KL}\big(q(\bar{\vz}_i) || p(\vz_i) \big)}_{\textrm{Dimension-wise KL}},
    \end{split}
\end{align}
where $p(n)$ denote the empirical data distribution.

\section{Experimental Setup}\label{app:sec:exp_setup}

\subsection{Disentanglement Study}

All hyperparameters for optimization and model architectures are listed in Table~\ref{app:table:exp_setup}. We compare our approach, GroupVAE, to four different models: $\beta$-VAE \cite{higgins2016beta}, AdaGVAE \cite{locatello2020weakly}, MLVAE \cite{bouchacourt2018multi} and GVAE \cite{hosoya2019group}. To fairly compare all models, we used the same architecture and optimization settings for all models and only varied the range of the regularization.  We ran five experiments for every hyperparameter set with different random seeds ($=[0, 1, 2, 3,  4]$). In total, we ran 240 experiments. Each experiment ran on GPU clusters consisting of Nvidia V100 or RTX 6000 for approximately 2-3 hours.

\paragraph{Datasets and group sampling.} We evaluated our approach on three datasets: 3D Cars \cite{reed2014learning}, 3D Shapes \cite{3dshapes18} and dSprites \cite{dsprites17}. All datasets contain images of size $64 \times 64$ with pixels normalized between 0 and 1. For training, given observations $\vx$  and groups $g_1,\ldots,g_m$, we sample uniformly $g$ from all groups and the observation $\vx'$ uniform from all observations which share the same group values as $\vx$.

\paragraph{Evaluating disentanglement.} In addition to comparing group disentanglement, we also used \pgls{mig} \cite{ChenLGD18} to compare the models' ability to disentangle all factors of variation. \cite{ChenLGD18} proposed MIG as an unbiased and hyperparameter-dependent evaluation metric to measure the mutual information between each ground truth factor and each dimension in the computed representation. The MIG is calculated as the average difference between the highest and second-highest normalized mutual information of each factor. The score is computed as
\begin{align}
    \textrm{MIG} = \frac{1}{K} \sum_{1}^{K} \frac{1}{H(v_k)} \big(I_n (\vz_{j^{(k)}}; v_k) - \max_{j \neq j^{(k)}} I_n(\vz_j;v_k) \big),
\end{align}
where $j^{(k)} = \argmax_{j} I_n(\vz_{j} ; v_k)$ and $K$ is the number of known factors.

\begin{table}[t]
    \centering
    \begin{subtable}[h]{0.4\textwidth}
        \centering
        \begin{tabular}{l r}
            \toprule
            \textbf{Parameters} & \textbf{Values} \\
            \toprule
            Batch size & 64 \\
            Latent dimension & 10\\
            Optimizer & Adam \\
            Adam: beta1 & 0.9\\
            Adam: beta2 & 0.999\\
            Adam: epsilon & 1$e$-8\\
            Adam: learning rate & 5$e$-4\\
            Training iterations & 300{,}000\\
            \bottomrule \\
        \end{tabular}
        \vspace{0.25cm}
        \caption{Common hyperparameters.}
    \end{subtable}
    \hfill
    \begin{subtable}[h]{0.59\textwidth}
        \centering
        \begin{tabular}{l l}
	\toprule
	\multicolumn{2}{l}{\textbf{Architecture}}\\
	\toprule
	$q_\phi(\vz|\vx)$ & Conv $32 \times 4 \times 4$ (Stride 2), ReLU act., \\
	& Conv $32 \times 4 \times 4$ (Stride 2), ReLU act.,\\
	 & Conv $64 \times 4 \times 4$ (Stride 2), ReLU act., \\
	 & Conv $64 \times 4 \times 4$ (Stride 2), ReLU act., \\
    & FC 256, ReLU act., FC 2 $\times$ 10\\
	$p_\theta(\vx|\vz)$ & FC 1024, ReLU act., Reshape (64, 4, 4),\\
	& TransposeConv $64 \times 4 \times 4$ (Stride 2), ReLU act.,\\
	& TransposeConv $32 \times 4 \times 4$ (Stride 2), ReLU act., \\
	& TransposeConv $32 \times 4 \times 4$ (Stride 2), ReLU act., \\
	& TransposeConv $3 \times 4 \times 4$ (Stride 2)\\
	\bottomrule
	\end{tabular}
	\caption{Common model architectures.}
    \end{subtable}
    \begin{subtable}[h]{0.5\textwidth}
    \centering
    \begin{tabular}{lcl}
    \toprule
    \textbf{Model} & \textbf{Parameter} & \textbf{Values}\\
    \toprule
    $\beta$-VAE \cite{higgins2016beta} & $\beta$ & $[1, 2, 4, 6, 8, 16]$\\
    AdaGVAE \cite{locatello2020weakly} & $\beta$ & $[1, 2, 4, 6, 8, 16]$\\
    MLVAE \cite{bouchacourt2018multi} & $\beta$ & $[1, 2, 4, 6, 8, 16]$\\
    GVAE \cite{hosoya2019group} & $\beta$ & $[1, 2, 4, 6, 8, 16]$\\
    GroupVAE & $\lambda$ & $[1, 2, 8, 16, 32, 64]$\\
    \end{tabular}
    \caption{Model hyperparameters.}
    \label{tab:my_label} 
    \end{subtable}
    \caption{\textbf{Experimental setup for the disentanglement study.} We list hyperparameters,  model architectures and hyperparamter common to the disentanglement study.}
    \label{app:table:exp_setup}
\end{table}
\subsection{Fairness}\label{app:groupvae:fair-exp-setting}
\begin{table}[t]
    \begin{subtable}[h]{\textwidth}
        \centering
        \begin{tabular}{l l}
    	    \toprule
    	    \multicolumn{2}{l}{\textbf{Architecture}}\\
    	    \toprule
    	    FFVAE discriminator & FC 1000, LeakyReLU(0.2) act., FC 1000, LeakyReLU(0.2) act., \\
    	    & FC 1000, LeakyReLU(0.2) act., FC 1000, LeakyReLU(0.2) act., \\
    	    & FC 1000, LeakyReLU(0.2) act., FC 2\\
    	    $f_{\textrm{MLP}}$ & FC 128, ReLU act., FC 128, ReLU act., FC 128, FC 2\\
    	    $f_{\textrm{CNN}}$ & Conv $1 \times 6 \times 5$, ReLU act., MaxPool\\
    	    & Conv $6 \times 16 \times 5$, ReLU act., MaxPool, FC 120,\\
    	    & ReLU act., FC 84, ReLU act., FC 2\\
    	    \bottomrule
	    \end{tabular}
	    \caption{Additional model architecture.}
    \end{subtable}\\
    \begin{subtable}[h]{0.5\textwidth}
        \centering
        \resizebox{\textwidth}{!}{
        \begin{tabular}{lcl}
            \toprule
            \textbf{Model} & \textbf{Parameter} & \textbf{Values}\\
            \toprule
            FFVAE & $\alpha$ & $[0, 1, 100, 300, 500, 1000]$\\
            & $\gamma$ & $[10, 20, 30, 40, 50, 100]$\\
            \bottomrule
        \end{tabular}
        }
    \caption{Additional model hyperparameter.}
    \end{subtable}
    \begin{subtable}[h]{0.5\textwidth}
        \centering
        \resizebox{\textwidth}{!}{
        \begin{tabular}{l l r}
            \toprule
            \textbf{Dataset} & \textbf{Parameters} & \textbf{Values} \\
            \toprule
            \multirow{2}{*}{CelebA} & Latent dimensions & \multirow{2}{*}{$[[3, 37], [40, 40]]$}\\
            & [sensitive, non-sensitive] & \\
            dSpritesUnfair & Latent dimensions & \multirow{2}{*}{$[[5, 5]]$}\\
            & [sensitive, non-sensitive] & \\
            \bottomrule
        \end{tabular}
        }
        \caption{Dataset-specific hyperparameters.}
    \end{subtable}
    \caption{\textbf{Experimental settings for fair classification.} We list hyperparameters of FFVAE and the MLP and CNN baselines.}\label{app:table:fair_exp_setup}
\end{table}
We ran five experiments for every hyperparameter set with different random seeds ($=[0, 1, 2, 3,  4]$). In total, we ran 550 experiments. Each experiment ran on GPU clusters consisting of Nvidia V100 or RTX 6000 for approximately 2-3 hours.
\paragraph{Models} For the fair classification experiments we used the same common hyperparameters and model architecture as in the disentanglement studies (Table~\ref{app:table:exp_setup} (a) and (b)) for GroupVAE, GVAE and MLVAE. In addition, we implemented two simple baselines, an \pgls{mlp} and a \pgls{cnn}. The architecture for these two models are described in Table~\ref{app:table:fair_exp_setup}. For the supervised fair classification, we implemented FFVAE \cite{creager2019flexibly} with the same encoder and decoder networks as in Table~\ref{app:table:exp_setup} (b) and the FFVAE discriminator as in Table~\ref{app:table:fair_exp_setup}. The baselines are trained with a cross-entropy loss between the logits of the network and the binary label ``HeavyMakeup''. We used different number of latent dimensions which is shown in Table~\ref{app:table:fair_exp_setup} (c).

\paragraph{Sensitive and non-sensitive latent variables.} Similar to the content and style disentanglement setup, we define two groups, sensitive and non-sensitive. GroupVAE can be optimized to learn from weakly supervised observations sharing either sensitive or non-sensitive group values. FFVAE~\cite{creager2019flexibly} can be seen as the supervised approach of learning sensitive and non-sensitive representations. FFVAE maximizes the ELBO objective (reconstruction loss and KL divergence between approximate posterior and prior). In addition, the objective regularizes the discriminative ability of the sensitive latent variable with $\alpha$ in a supervised manner (\textit{how well can the model classify sensitive labels from sensitive latent variable?}) and the disentanglement with $\gamma$ (\textit{how well is the sensitive latent variable disentangled from the non-sensitive latent variable?}).

\paragraph{Datasets.} For comparability with FFVAE~\cite{creager2019flexibly}, we used similar dataset settings for CelebA~\cite{li2018deep} and dSpritesUnfair. Both datasets contain images with pixels normalized between 0 and 1. We used the pre-defined train, validation, and test split of CelebA~\cite{li2018deep}, whereas in dSpritesUnfair we use a random split of 80\% train, 5\% validation, and 15\% test.

\paragraph{dSpritesUnfair.} dSpritesUnfair is a modified version of dSprites~\cite{dsprites17}. The two components are the binarization of the factors of variation and biased sampling. dSprites contains images which are described by five factors of variation. We binarized the factors of variations following these criterion~\cite{creager2019flexibly}:
\begin{itemize}
    \item Shape $\ge 1$
    \item Scale $\ge 3$
    \item Rotation $\ge 20$
    \item X-position $\ge 16$
    \item Y-position $\ge 16$
\end{itemize}
Similar to \cite{trauble2020independence}, we enforce correlations between shape and x-position through a biased sampling. In the training set, we sample these two factors from a joint distribution
\begin{align}
    p(s, x) \propto \textrm{exp}\big(-\frac{(s - x)^2}{2\sigma^2}\big),
\end{align}
where $\sigma$ determines the strength of the.correlation and is set to $\sigma=0.2$ in our experiments. The smaller $\sigma$, the higher the correlation between the two factors.
\paragraph{Model selection.}
As shown in~\cite{creager2019flexibly}, there is a trade-off between classification accuracy and demographic parity. Thus, model selection based on only one of these metrics compromises the other. We propose to use the difference between the two metrics as a way to do model selection.
We coin this metric FairGap (FG) and define it as
\begin{align}
    FG = \underbrace{\mathbb{E}[\bar{y} = y]}_{\textrm{Accuracy}} - \frac{1}{|a|} \sum_{a} \underbrace{|\mathbb{E}[\bar{y} = 1|a = 1] - \mathbb{E}[\bar{y} = 1|a = 0]|}_{\textrm{demographic parity}}.
\end{align}
FG is high if accuracy is high and the average demographic is low, resulting in a fair classifier. We select the model on the test set of CelebA and dSprites based on the FG of the validation set.

\end{document}